\pdfoutput=1
\documentclass[journal,twoside,web]{ieeecolor}
\usepackage{generic}
\usepackage{cite}
\usepackage{amsmath,amssymb,amsfonts}
\usepackage{algorithm}
\usepackage{algpseudocode}
\usepackage{array}
\usepackage{caption}
\usepackage[font=normalsize,labelfont=sf,textfont=sf]{subcaption}
\usepackage{stfloats}
\usepackage{graphicx}
\usepackage{url}
\usepackage{verbatim}
\usepackage{booktabs}
\usepackage{multirow}
\usepackage[table,dvipsnames, x11names]{xcolor}
\usepackage{hyperref}
\usepackage{tabularx}
\hypersetup{hidelinks}
\usepackage{adjustbox}
\usepackage{textcomp}

\def\BibTeX{{\rm B\kern-.05em{\sc i\kern-.025em b}\kern-.08em
    T\kern-.1667em\lower.7ex\hbox{E}\kern-.125emX}}

\makeatletter
\renewcommand*\env@matrix[1][\arraystretch]{%
  \edef\arraystretch{#1}%
  \hskip -\arraycolsep
  \let\@ifnextchar\new@ifnextchar
  \array{*\c@MaxMatrixCols c}}
\makeatother

\begin{document}
\title{Omni-modal decomposition autoencoders learn full-stack wearable disentangled representations}

\author{Ioannis N. Ziogas, \IEEEmembership{Student Member, IEEE}, Ensieh Khazaei, Bilal Taha, Aamna Al Shehhi, Ahsan H. Khandoker, \IEEEmembership{Senior Member, IEEE}, Leontios J. Hadjileontiadis, \IEEEmembership{Senior Member, IEEE}, Dimitrios Hatzinakos, \IEEEmembership{Fellow, IEEE} 
\thanks{Manuscript received \today. This research was funded by the Healthcare Engineering Innovation Group (HEIG), Khalifa University of Science and Technology (KU-HEIG). }
\thanks{Ioannis N. Ziogas is with the Department of Biomedical Engineering and Biotechnology, Khalifa University, 127788 Abu Dhabi, UAE. Part of this work was conducted during his time as a Visiting Graduate Researcher at the Edward S. Rogers Sr. Department of Electrical and Computer Engineering, University of Toronto, Toronto, ON M5S 1A1, Canada (e-mail: ioannis.ziogas@ku.ac.ae).}
\thanks{Ensieh Khazaei is with the Edward S. Rogers Sr. Department of Electrical and Computer Engineering, University of Toronto,
Toronto, ON M5S 1A1, Canada.}
\thanks{Bilal Taha is with the MIT Media Lab and Edward S. Rogers Sr. Department of Electrical and Computer Engineering, University of Toronto.}
\thanks{Aamna Al Shehhi is with the Department of Electrical Engineering, Khalifa University, Abu Dhabi, UAE.}
\thanks{Ahsan H. Khandoker is with the Department of Biomedical Engineering and Biotechnology, Khalifa University, 127788 Abu Dhabi, UAE.}
\thanks{Leontios J. Hadjileontiadis is with the Department of Electrical and Computer Engineering, Aristotle University of Thessaloniki, GR 54124 Thessaloniki, Greece, and with the Department of Biomedical Engineering and Biotechnology, Khalifa University, 127788 Abu Dhabi, UAE (e-mail: leontios\{@auth.gr, .hadjileontiadis@ku.ac.ae)\}.}
\thanks{Dimitrios Hatzinakos is with the Edward S. Rogers Sr. Department of Electrical and Computer Engineering, University of Toronto,
Toronto, ON M5S 1A1, Canada (e-mail: dimitris@comm.utoronto.ca).
}
\thanks{Code:\textit{\url{https://github.com/GiannisZgs/OmniDecVAEs}}}
}

\IEEEaftertitletext{\vspace{-3\baselineskip}}
\maketitle
\begin{abstract}
Learning disentangled representations is key for general-purpose multi-modal wearable models, yet existing approaches rarely combine classification, disentanglement, fusion, and generation of heterogeneous multi-modal time series in a single model. We introduce Omni-modal Variational Decomposition Autoencoders (OmniDecVAEs), a framework that learns multi-purpose representations from arbitrarily many modalities with a single shared encoder-decoder. OmniDecVAEs extend DecVAEs by learning time-frequency latent subspaces per modality through a multi-view self-supervised decomposition loss and a component-wise conditional decoder. \textcolor{black}{On the omni-modal human activity recognition (HAR) HARWE dataset (up to thirty channels) and the public UCI HAR and WESAD datasets, OmniDecVAEs learn the most disentangled representations for unseen subjects (DCI disentanglement up to 3.0 percentage points (pp) higher). On HARWE, they improve subject-dependent HAR and identity recognition by 1.1 pp and 5.8 pp, and outperform supervised end-to-end models by +31.6 pp and +22.3 pp on unseen subjects of multi-device datasets. Moreover, their generative capacity improves reconstruction mean squared error over multi-modal VAEs (34-74$\%$ improvement) and their synthetic data improve affect recognition on unseen subjects (2 pp).} \textcolor{black}{With 4.1M parameters, invariant to the number of modalities (21$\times$ fewer than a multi-branch VAE), OmniDecVAEs are scalable, single-model full-stack learners for wearable representations}. 
\end{abstract}

\section{Introduction}
\label{sec:introduction}
\IEEEPARstart{W}{earable} sensing systems are increasingly generating large-scale, heterogeneous multi-modal data, requiring robust and expressive representation learning for applications in healthcare, human-computer interaction, and biometric security \cite{kang2023kemophone}. Recent advances in sensing hardware enable the integration of diverse physiological and behavioral modalities within unified wearable platforms, resulting in increasingly complex multi-modal data representations. This evolution motivates a transition from conventional multi-modality to \textit{omni-modality}, defined as the unification of arbitrarily many heterogeneous data streams into universal and expressive representations through a single model \cite{Zhang2025ScalingModalities}.

\textcolor{black}{Supervised multi-modal approaches, often based on transformers~\cite{Minor2026AData} or state-space models~\cite{Avramidis2024ScalingModels}, require large amounts of labeled data. Fusing heterogeneous streams such as inertial, physiological, and behavioral data~\cite{Esmaeilzehi2024HARWE:Environments} requires per-modality design, even for closely related modalities, and even more so when combining disparate modalities~\cite{Fan2026CMD3:Fusion}, such as video and inertial data \cite{He2026Causal-InspiredRecognition}, or biosignals and audio \cite{Wu2024Transformer-BasedRecognition}. Self-supervised learning (SSL) has emerged as a powerful alternative, enabling the learning of transferable representations with minimal annotations \cite{balestriero2023} through contrastive~\cite{Xiao2025CapMatch:Recognition}, or reconstruction-based objectives ~\cite{Liu2024RPPG-MAE:Measurements}. However, extending SSL to multi-modal wearable settings requires architectural modifications~\cite{Zhou2025MSMFT:Recognition}, graph-based formulations~\cite{Ziogas2025Self-SupervisedRecognition}, or tailored objectives~\cite{He2026Causal-InspiredRecognition}. Moreover, existing approaches focus on task-specific performance, without addressing disentanglement and generation. Autoencoder (AE)-based methods \cite{Han2020DisentangledExtraction, Su2025EMGRepresentations, He2026Causal-InspiredRecognition} and generative adversarial networks (GANs) \cite{Dissanayake2023GeneralizedTransfer} promote disentangled representations, which enable applications such as biometric privacy and identity-aware modeling \cite{Su2025EMGRepresentations}, but typically rely on modality-specific encoders or decoders \cite{Su2025EMGRepresentations,Zhang2025PPGNetworks}. To the best of our knowledge, no existing approach jointly addresses scalable multi-modal fusion, disentanglement, and generation within a single framework.}

\textcolor{black}{Here, we introduce OmniDecVAE, a self-supervised, structure-aware framework for omni-modal wearable computing that learns \emph{full-stack} representations by treating individual modality channels as components of a unified system through an adversarial contrastive objective inspired by signal decomposition~\cite{Ziogas2026VariationalRepresentations}. All inputs, up to thirty modality channels in our experiments, are transformed into the time-frequency (TF) domain; a shared convolutional encoder learns modality-specific latent subspaces, and a shared fully-connected (FC) decoder, conditioned on the component, reconstructs and generates multi-modal signals through the variational formulation.}

We evaluate the representations with conventional supervised classifiers \textcolor{black}{and standard disentanglement metrics} on HARWE~\cite{Esmaeilzehi2024HARWE:Environments}, a large-scale multi-modal dataset for human activity and identity recognition \textcolor{black}{with up to thirty modality channels, and on the UCI HAR~\cite{anguita2013public} and WESAD~\cite{schmidt2018introducing} datasets.
Compared with supervised transformer-based fusion~\cite{Esmaeilzehi2024HARWE:Environments}, VAE-based models~\cite{Kingma2014Auto-EncodingBayes},\cite{Higgins2017Beta-VAE:Framework} and eigen-projections ICA~\cite{hyvarinen2001independent} and PCA~\cite{Greenacre2022PrincipalAnalysis}, OmniDecVAE learns the most disentangled representations for unseen subjects on all three datasets, reconstructs more faithfully than multi-modal VAEs, is the most accurate model on HARWE for known subjects, and is far more accurate than supervised end-to-end models for unseen subjects on HARWE and WESAD. These properties are relevant to wearable and clinical applications, such as adapting to fast changing sensing environments, identity-aware modeling and anonymization, and reconstruction of missing data.  
}

In this work, we make the following contributions:
\begin{itemize}
    \item We propose a multi-modal SSL objective that extends variational decomposition autoencoding to fuse arbitrarily many modalities in a structured latent space, with a modality-agnostic architecture.
    \item We introduce a shared encoder-decoder architecture, that scales to many modalities without relying on transformer-based designs, and generates multi-modal TF data.
    \item We show on three datasets that the learned representations are more disentangled than those of multi-modal VAEs, supervised transformers and eigen-projections, while remaining competitive in downstream recognition, and having more faithful reconstruction than multi-modal VAEs.
\end{itemize}

\section{Related Work} 
\label{sec:relwork}

\subsection{Multi-modal Fusion in Wearable Computing}

Concatenation-based fusion is a common approach that aggregates modalities at the input (early fusion) or in the latent space (late fusion), with a separate feature extractor per modality, e.g., convolutional branches on TF representations~\cite{SriramKumar2024DeepMethods}. Transformer-based models model intra- and inter-model dependencies through attention, but still depend on modality-specific deep feature extractors~\cite{Wu2024Transformer-BasedRecognition,Dissanayake2022SigRep:Learning}, whereas contrastive alignment reduces the latent distance between the modalities of a sample with separate encoders~\cite{Nguyen2024VirtualRecognition}. Overall, existing fusion methods do not scale to an arbitrary number of modalities without architectural modifications.

\subsection{Self-supervised Learning in Multi-modal Wearable Computing}

In wearable sensing, SSL can learn representations from unlabeled data through surrogate objectives such as reconstruction or contrastive learning~\cite{he2022MAE}. In multi-modal wearable settings, SSL facilitates fusion through augmentation-prediction tasks~\cite{Dissanayake2022SigRep:Learning,Wu2024Transformer-BasedRecognition}, temporal neighborhood contrastive pairs (TCLHAR~\cite{Chen2025TemporalApproach}), or reconstruction across modality branches~\cite{He2026Causal-InspiredRecognition}. SSL approaches that relate the modalities directly through cross-modal modeling, maximize the agreement between the embeddings of different modalities at the same time step (COCOA~\cite{deldari2022cocoa}), mask intermediate modality embeddings and combine them through a cross-modal aggregator (CroSSL~\cite{deldari2024crossl}), or factorize each modality's representation into shared and private features under orthogonality constraints (FOCAL~\cite{liu2023focal}). However, these objectives use modality-specific encoders and thus handle cross-modal fusion through architectural design, and none models signal generation.

\subsection{Multi-modal Generation through Disentangled Representations in Wearable Computing} 

Disentangled representation learning separates the factors of variation in high-dimensional data \cite{carbonneau2022dis_metrics}, such as identity from activity in wearable sensing~\cite{Su2025EMGRepresentations}. Multi-modal VAEs infer a joint posterior from modality-specific encoders, combined as a product~\cite{wu2018multimodal}, a mixture~\cite{shi2019variational}, or a mixture of products of experts~\cite{sutter2021generalized}, so their size grows with the number of modalities; our MMVAE baseline has modality-dedicated branches with the capacity of OmniDecVAE, does not model a joint posterior, and is thus distinct from~\cite{sutter2021generalized}. In wearable computing, multi-branch AEs combine reconstruction, adversarial subject invariance and classification without assessing generation~\cite{Han2021UniversalAutoencoders}, whereas in electromyography(EMG)-based systems~\cite{Su2025EMGRepresentations} separate encoders disentangle identity. Approaches such as GANs have been used to transfer between related modalities such as ECG and PCG~\cite{Dissanayake2023GeneralizedTransfer, Karimi2025BidirectionalPCG}, but without evaluating downstream utility of learned representations.

%
\begin{figure*}[ht!]
    \includegraphics[height=8.5cm,width=0.85\textwidth]{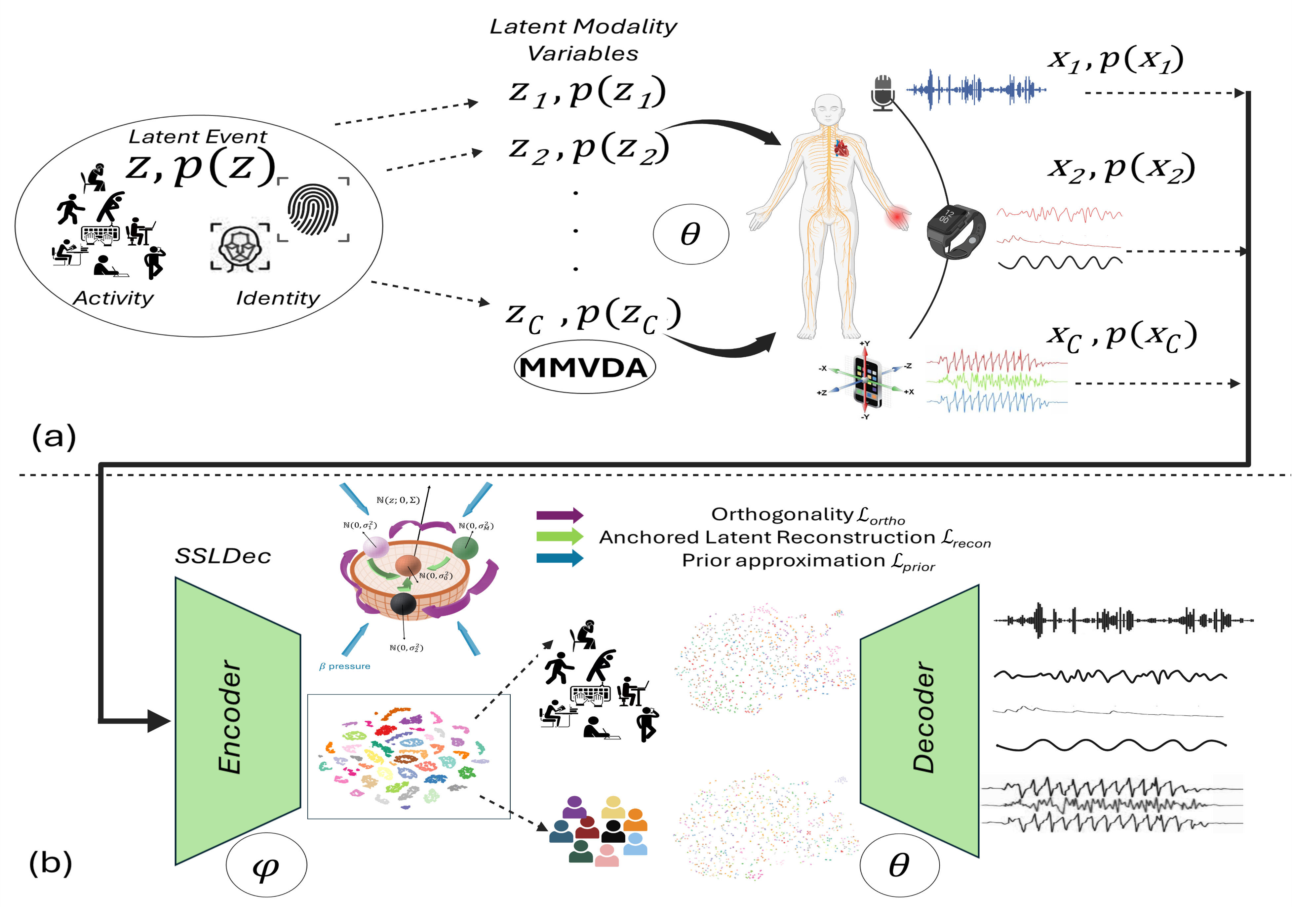}
    \centering
    \caption{Overview of OmniDecVAEs and MMVDA, the proposed omni-modal full-stack wearable disentangled representation learning framework. (a) Three-step generative process $\theta$: a latent event associated with activity and identity variables is generated, decomposed into \textcolor{black}{$C$} latent modality variables, and expressed through wearable sensor measurements across physiological, inertial, and acoustic domains. (b) MMVDA instantiated through OmniDecVAE aims to learn, via the inference model $\phi$, a reverse mapping from inputs $\mathcal{X}$ to a modality-separated latent space $\mathcal{Z}$, ultimately recovering the underlying activity- and identity-conditioned latent event. The interplay between the encoder $\phi$, the SSLDec structure-aware latent objective, and the decoder $\theta$ enables effective disentanglement, supports downstream classification tasks, and facilitates data synthesis. }
    \label{fig:graphical_abstract}
\end{figure*}

\section{Proposed Method} 
\label{methods}
\subsection{Problem Statement}
\label{problem_statement}

Let \textcolor{black}{$\mathcal{X}=\{\mathbf{x}_1,\mathbf{x}_2,\dots,\mathbf{x}_N\}$} denote a multivariate time-series dataset with $N$ i.i.d. observations, \textcolor{black}{recorded by $M$ modalities}. Each observation \textcolor{black}{ $x_n \in \mathbb{R}^K$ contains the channels of all sensors, and a channel may be further split into oscillatory components (OCs) by a decomposition model, in complex modalities such as audio (see Section~\ref{experiments}). We define every channel that is not split and every OC, as a \emph{component} $\mathbf{x}^{(i)}$ and denote their number by $C$; for example, in the full HARWE~\cite{Esmaeilzehi2024HARWE:Environments} setting, $C=30$}. Each observation \textcolor{black}{is} generated by a three-step generative process $\theta$, involving a global latent variable $\mathbf{z}$ and a set of \textcolor{black}{component} latent variables \textcolor{black}{$\{\mathbf{z}_1, \mathbf{z}_2, ..., \mathbf{z}_C\}$}, \textcolor{black}{each representing a distinct view of the \emph{sensing event} $\mathbf{z}$ through one component (e.g., physiological, inertial, and acoustic, see Fig.~\ref{fig:graphical_abstract}a)}. \textcolor{black}{Each $\mathbf{z}_i$ takes values in its own subspace $\mathcal{Z}_i$, and the latent space is their sum $\mathcal{Z} = \mathcal{Z}_1\oplus\cdots\oplus\mathcal{Z}_C$}. Under the assumption \textcolor{black}{that each $\mathbf{z}_i$ identifies with its embedding in $\mathcal{Z}$}, the relationships between latent variables are defined as:

\textcolor{black}{
\begin{align}
\langle &\mathbf{z}_i,\mathbf{z}_j\rangle = 0, \quad \forall\, i\neq j, \label{eq_ortho_generative_space}\\
\mathbf{z} &= \textstyle\sum_{i=1}^{C}\mathbf{z}_i = [\mathbf{z}_1;\mathbf{z}_2;\dots;\mathbf{z}_C]
\label{eq_recon_generative_space}
\end{align}
}

\noindent
where $\langle \cdot, \cdot \rangle$ denotes the dot product in $\mathcal{Z}$ and $[\,\cdot\,;\cdot\,]$ concatenation. \textcolor{black}{In this idealized setting, the component latents occupy orthogonal subspaces and their aggregation fully reconstructs the latent event $\mathbf{z}$, so that additional sensing modalities can yield progressively richer representations of $\mathbf{z}$, thus naturally motivating an \emph{omni-modal} setting}. \textcolor{black}{Properties of $\mathbf{z}$, such as activity and identity, characterize $\mathbf{z}$ as a whole, and this formulation does not assume that a single $\mathbf{z}_i$ encodes them; their distribution across subspaces is assessed empirically (Section~\ref{results})}.

Unlike VDA~\cite{Ziogas2026VariationalRepresentations} where a decomposition model estimates the components of a univariate signal, here the observation is already decomposed into modalities and channels; the objective thus shifts to learning a \emph{composition} model that expresses $\mathbf{z}$ across the components. This introduces three new elements in MMVDA, (i) the construction of an anchor, since no single input contains all information (Section~\ref{anchor_methods}), (ii) the omni-modal interactions of heterogeneous components are not uniform (Section~\ref{omni_modal_contrasting}), (iii) the generative model must respect subspace orthogonality and infer components separately (Sections~\ref{mmvda_objective} and \ref{decoder_theory}).

\noindent\textbf{Method Overview.}
The proposed omni-modal framework outlined across the following sections can be summarized as follows: 
(1) \emph{Input:} given a multi-modal observation \textcolor{black}{$\mathbf{x}$ with components $\{\mathbf{x}^{(1)}, \dots, \mathbf{x}^{(C)}$}, the anchor \textcolor{black}{$\mathbf{x}^{(0)} = A(\mathbf{x})$} is constructed \textcolor{black}{(Eq.~\eqref{eq_anchor})}. 
(2) \emph{Encoding:}\textcolor{black}{components} and the anchor are fed to a shared encoder $f(\cdot)$ to obtain \textcolor{black}{embeddings} $\mathbf{h}_i$ and $\mathbf{h}_0$. 
(3) \emph{Latent projection:} \textcolor{black}{the embeddings} are mapped to the latent variables \textcolor{black}{$\mathbf{z}_i \sim q_{\phi}(\mathbf{z}_i \mid \mathbf{x}^{(i)}), i \in\{1,C\}$}. 
(4) \emph{Self-supervised decomposition:} the objective of Eq.~\eqref{eq_decomp_loss_compute} enforces (i) alignment of each $\mathbf{h}_i$ with the anchor $\textbf{h}_0$ via $\mathcal{L}_{\text{recon}}$ and (ii) orthogonality via $\mathcal{L}_{\text{ortho}}$, using omni-modal weights (Eq.~\eqref{eq_asymm_weights}). 
(5) \emph{Latent aggregation:} the subspaces are combined into the representation $\mathbf{\tilde{z}} = [\mathbf{\tilde{z}}_1;\mathbf{\tilde{z}}_2;\dots;\mathbf{\tilde{z}}_C]$. 
(6) \emph{Decoding:} the latent variables \textcolor{black}{$z_i, i \in\{0,C\}$} are sampled via reparameterization (Eq.~\eqref{eq_reparam_trick}) and \textcolor{black}{decoded by} the conditional decoder \textcolor{black}{$p_{\theta}(\mathbf{x}^{(i)} \mid \mathbf{z}_i, \mathbf{c}_i)$.} 
(7) \emph{Optimization:} the model is trained with the \textcolor{black}{multi-modal VDA objective, without (Eq.~\eqref{eq_mmvda}) or with (Eq.~\eqref{eq_mmvda_dec}) the decoder.}

\subsection{Multi-modal Variational Decomposition Autoencoding}
\label{mmvda_theory}
\textcolor{black}{MMVDA instantiates the process $\theta$ of Fig.~\ref{fig:graphical_abstract}a: (i) the latent components $\{\mathbf{z}_1, \mathbf{z}_2, ..., \mathbf{z}_C\}$ are sampled independently from their priors, (ii) they compose $\mathbf{z}^{n}$ (Eq.~\eqref{eq_recon_generative_space}), and (iii) each component signal is generated from its own latent, with the final observation being their superposition:}

\textcolor{black}{
\begin{subequations}
\label{eq_joint_probability}
\begin{align}
p_\theta(\mathcal{X},\mathbf{z}^{1:N}) &= \textstyle\prod_{n=1}^{N} p_\theta(\mathbf{x}_n\mid\mathbf{z}^n_1,\dots,\mathbf{z}^n_C)\,p_\theta(\mathbf{z}^n),\\
p_\theta(\mathbf{z}^n) &= \textstyle\prod_{i=1}^{C}p_\theta(\mathbf{z}^n_i),
\end{align}
\end{subequations}
\begin{equation}
\label{eq_gen_model_dists}
\begin{aligned}
p_\theta(\mathbf{x}\mid\mathbf{z}_1,\dots,\mathbf{z}_C) &= \mathcal{N}\!\big(\mathbf{x}\,\big|\,\textstyle\sum_{i=1}^{C} g_i(\mathbf{z}_i),\,\boldsymbol{\Sigma}_x\big),\\
p_\theta(\mathbf{z}_i) &= \mathcal{N}\!\big(\mathbf{z}_i\,\big|\,\mathbf{0},\,\mathbf{I}\big),\quad i\in\{1,\dots,C\},
\end{aligned}
\end{equation}
}
\noindent 
\textcolor{black}{where $g_i$ is the generative model that maps $\mathbf{z}_i$ to the observation subspace $\mathcal{V}_i \subset \mathbb{R}^K$ of component $i$, among mutually orthogonal subspaces $\mathcal{V}_1 \oplus \cdots \oplus \mathcal{V}_C = \mathbb{R}^K$ (the input domain counterparts of Eqs.~\eqref{eq_ortho_generative_space} and \eqref{eq_recon_generative_space}), and $\mathbf{\Sigma}_x = \sum_{i=1}^C \sigma^2_i \mathbf{P}_i$ with $\mathbf{P}_i$ the orthogonal projection onto $\mathcal{V}_i$, gives each component its own noise level $\sigma^2_i$. Under Eqs.~\eqref{eq_joint_probability},~\eqref{eq_gen_model_dists}, the posterior factorizes over components, with each factor depending on $\mathbf{x}$ only through $\mathbf{x}^{(i)} = \mathbf{P}_i \mathbf{x}$ (Proposition 1 in VDA~\cite{Ziogas2026VariationalRepresentations}). We note that for sensor channels the projection $\mathbf{P}_i$ is an exact channel selection; approximation through a decomposition model, as done in VDA~\cite{Ziogas2026VariationalRepresentations}, is only used for OCs. Following, the recognition model~\cite{Kingma2014Auto-EncodingBayes} is given the same structure:}
\textcolor{black}{
\begin{equation}
\label{eq_posterior_model}
\begin{aligned}
q_\phi(\mathbf{z}_1,\dots,\mathbf{z}_C\mid\mathbf{x})&=\textstyle\prod_{i=1}^{C}q_\phi\big(\mathbf{z}_i\mid\mathbf{x}^{(i)}\big),\\
q_\phi\big(\mathbf{z}_i\mid\mathbf{x}^{(i)}\big)&=\mathcal{N}\!\big(\mathbf{z}_i\,\big|\,f_{\mu_{z_i}}(\mathbf{x}^{(i)}),\,\mathrm{diag}\,f_{\sigma^2_{z_i}}(\mathbf{x}^{(i)})\big),
\end{aligned}
\end{equation}
}
\noindent
\textcolor{black}{where $f_{\mu_{z_i}}$ and $f_{\sigma^2_{z_i}}$ share all parameters except their final linear projections (see Section~\ref{decoder_theory}). The encoding of shared factors such as activity and identity in the modality embeddings, which the recognition model of Eq.~\eqref{eq_posterior_model} does not model, is carried out by the anchor (Section~\ref{anchor_methods}).}


\subsection{Multi-modal Multiple Views Self-supervised Latent Decomposition Contrastive Learning}
\label{multimodal_ssldec}

The recognition model of Eq.~\eqref{eq_posterior_model} factorizes over components, and hence the embedding space of an encoder is not restricted to satisfy Eqs.~\eqref{eq_ortho_generative_space},\eqref{eq_recon_generative_space}. Therefore,
we employ the self-supervised latent decomposition (SSLDec) loss of \cite{Ziogas2026VariationalRepresentations} that imposes Eqs.~\eqref{eq_ortho_generative_space}, \eqref{eq_recon_generative_space} in an intermediate embedding space $\mathcal{H}$. Let $\mathbf{h}_{n,i,f}$ and $\mathbf{h}_{n,f}$ be the embeddings of frame $f$ of component $i$ and of the anchor $\mathbf{x}_n^{(0)}$ (see Section~\ref{anchor_methods}) of observation $n$, respectively, obtained through the shared encoder and projector $f_H\circ f_W$ (Section~\ref{omnidecvae_methods}), and $r_{n,i,f}, p_{n,i,f}$ their softmax normalized embeddings over the $d$ dimensions of $\mathcal{H}$. As in~\cite{Ziogas2026VariationalRepresentations}, SSLDec's core mechanism is based on the Jensen-Shannon divergences $D_{JS}$ (given in nats, at most $\ln 2$), summed and clipped over the set of frames $\mathcal{F}_n$ of an observation, as:  

\textcolor{black}{
\begin{equation}
\label{eq_jsd}
\begin{aligned}
s^{pos}_{n,i}&=\min\Big\{1,\textstyle\sum_{f\in\mathcal{F}_n}D_{JS}\big(\mathbf{r}_{n,i,f}\|\mathbf{p}_{n,f}\big)\Big\},\\
s^{neg}_{n,ij}&=\min\Big\{1,\textstyle\sum_{f\in\mathcal{F}_n}D_{JS}\big(\mathbf{r}_{n,i,f}\|\mathbf{r}_{n,j,f}\big)\Big\},
\end{aligned}
\end{equation}
}
\noindent
\textcolor{black}{For a mini-batch of $B$ observations, the objective is formulated as:}

\textcolor{black}{
\begin{equation}
\label{eq_decomp_loss_compute}
\begin{aligned}
\mathcal{L}_{recon}&=\tfrac{1}{B}\textstyle\sum_{n=1}^{B}\sum_{i=1}^{C} w^{i}_{pos}\big[-\ln\big(1-s^{pos}_{n,i}\big)\big],\\
\mathcal{L}_{ortho}&=\tfrac{1}{B}\textstyle\sum_{n=1}^{B}\sum_{i<j} w^{(i,j)}_{neg}\big[-\ln s^{neg}_{n,ij}\big],\\
\mathcal{L}_{SSLDec}&=\mathcal{L}_{recon}+\mathcal{L}_{ortho},
\end{aligned}
\end{equation}
}
\noindent
here $w^i_{pos}$ and $w^{(i,j)}_{neg}$ weight positive and negative interactions (Section~\ref{omni_modal_contrasting}). $\mathcal{L}_{SSLDec}$ is a multi-view contrastive objective with an adversarial structure: $\mathcal{L}_{recon}$ aligns each component embedding with the anchor embedding, whereas $\mathcal{L}_{ortho}$ separates component embeddings; their competition is stabilized by the bounded $D_{JS}$ and the divergence of $L_{ortho}$ under representation collapse~\cite{Ziogas2026VariationalRepresentations}.

\textcolor{black}{For a single frame, equal weights $w^i_{pos}$ per component, and $d\geq C$ (in our experiments, $d=192$), Proposition~2 in VDA~\cite{Ziogas2026VariationalRepresentations} states that (a) $\mathcal{L}_{ortho}$ is minimized if and only if the components' normalized embeddings occupy pairwise disjoint sets of dimensions, (b) $\mathcal{L}_{recon}$ is minimal if the anchor is an even mixture of the above components $\mathbf{p}=\frac{1}{C}\sum_i\mathbf{r_i}$, (c) $\mathcal{L}_{recon},\mathcal{L}_{ortho}$ cannot vanish together. This equilibrium realizes Eqs.~\eqref{eq_ortho_generative_space}, \eqref{eq_recon_generative_space} in the intermediate space $\mathcal{H}$, but does not imply statistical independence of the latent subspaces or correspondence to semantic factors.}

\subsection{\textcolor{black}{Multi-modal Anchor}}
\label{anchor_methods}

In VDA~\cite{Ziogas2026VariationalRepresentations} the anchor is the unimodal input from which the components are decomposed; in the multi-modal setting, channels and OCs have different sampling rates and units. Hence, MMVDA requires a composite input as the anchor of $\mathcal{L}_{SSLDec}$. To that end, we construct the anchor as a fixed, non-learnable aggregation map of the observation $\mathbf{x}$:

\begin{equation}
\mathbf{x}^{(0)}=A(\mathbf{x}),
\label{eq_anchor}
\end{equation}

\noindent
embedded by the same recognition model. The anchor is not part of the generative model (Eqs.~\eqref{eq_joint_probability}-\eqref{eq_posterior_model}), and acts as the reference embedding of $\mathcal{L}_{recon}$, which drives its normalized embedding towards the even mixture of the component embeddings (Prop.~2(b) in~\cite{Ziogas2026VariationalRepresentations}), which is similar to mixture-of-experts fusion in multi-modal VAEs~\cite{shi2019variational}, or towards strongly weighted components for unequal weights (Section~\ref{omni_modal_contrasting}). Since the generative and posterior models factorize over components, the anchor latent $\mathbf{\tilde{z}}_0$ is the only subspace of the representation $\mathbf{\tilde{z}}_0 = [\mathbf{\tilde{z}}_0;\mathbf{\tilde{z}}_1;\dots;\mathbf{\tilde{z}}_C]$ with information from all modalities. The equilibrium of $\mathcal{L}_{recon}$ requires the anchor to contain information for every component~\cite{Ziogas2026VariationalRepresentations}, which motivates aggregating all channels in the anchor construction. We use the sum of standardized channels as the default, as this setting brings different modalities to a common scale, and we ablate $A(\cdot)$ in Section~\ref{results}. It should be noted that a component-wise decoder also relates components to the whole observation, but in the input superposition domain (Eq.~\eqref{eq_gen_model_dists}); we therefore expect the anchor to matter less in the presence of a decoder (Section~\ref{results}).


\subsection{\textcolor{black}{Increase and Asymmetry of Contrastive Pairs}}
\label{omni_modal_contrasting}

In wearable systems, related modalities (e.g. accelerometer (ACC), and gyroscope (GYR), electrodermal activity (EDA) and blood volume pulse (BVP)) and channels of the same sensor (e.g., tri-axial ACC) share stronger relationships than unrelated channels, and influence the anchor non-uniformly. SSLDec assumes uniform interactions, scaling the objective with weights that sum to one, $\sum_i w_{pos}^i = \sum_{i<j}w_{neg}^{(i,j)}=1$~\cite{Ziogas2026VariationalRepresentations}. In the current omni-modal setting, we weight the interactions asymmetrically: a set $\mathcal{S}$ of $N_{\mathcal{S}}$ components receives stronger positive weights than the remaining $N_w=C-N_\mathcal{S}$, and a set $\mathcal{W}$ of component pairs receives weaker negative weights, controlled by the imbalance factors $IF_{pos},IF_{neg}$:

\textcolor{black}{
\begin{equation}
\begin{aligned}
w^{i}_{pos}&=\frac{1}{C}\begin{cases}1+IF_{pos}\frac{N_w}{N_s}, & i\in\mathcal{S},\\[2pt] 1-IF_{pos}, & i\notin\mathcal{S},\end{cases}\\
w^{(i,j)}_{neg}&=\binom{C}{2}^{-1}\begin{cases}1-IF_{neg}, & (i,j)\in\mathcal{W},\\[2pt] 1+IF_{neg}\frac{|\mathcal{W}|}{\binom{C}{2}-|\mathcal{W}|}, & \text{otherwise},\end{cases}
\end{aligned}
\label{eq_asymm_weights}
\end{equation}
}
\noindent
where $0\leq IF < 1$, and $IF=0$, gives uniform weights; we set $IF_{pos}=0.25$ and $IF_{neg}=0.4$ and examine both in Section~\ref{results}. Larger positive weights drive the anchor towards modalities of higher complexity or dimensionality, such as audio in HARWE~\cite{Esmaeilzehi2024HARWE:Environments} or EMG in WESAD~\cite{schmidt2018introducing}; for comparable modalities, e.g., inertial only as in UCI HAR~\cite{anguita2013public}, $\mathcal{S}$ is empty and $IF_{pos}=0$. Furthermore, weaker negative weights are assigned to channels expected to be similar, such as physiological readings of the same device (e.g. BVP, EDA, TEMP) and channels of the same sensor (e.g., X-axes of ACC and GYR, three axes of ACC), and stronger weights to dissimilar modalities. Because the weights keep their normalized sum property, MMVDA can scale to arbitrarily many channels, modalities and OCs; notably, the negative weights do not move the minimizer of $\mathcal{L}_{ortho}$ but only redistribute the separation pressure across pairs~\cite{Ziogas2026VariationalRepresentations}.


\subsection{\textcolor{black}{MMVDA Objective and Component-wise Decoder}}
\label{mmvda_objective}
We instantiate MMVDA by parameterizing the generative and inference models $p_\theta$ and $q_\phi$ \textcolor{black}{as} neural networks \textcolor{black}{The MMVDA objective without a decoder, is ${L}_{SSLDec}$ and a prior term on the component subspaces, as follows:}

\textcolor{black}{
\begin{equation}
\begin{aligned}
\mathcal{L}_{MMVDA}&=\mathcal{L}_{recon}+\mathcal{L}_{ortho}+\beta\,\mathcal{L}_{prior},\\
\mathcal{L}_{prior}=\tfrac{\gamma}{C}\textstyle\sum_{n=1}^{B}&\sum_{i=1}^{C}D_{KL}\big(q_\phi(\mathbf{z}_i\mid\mathbf{x}^{(i)}_n)\,\|\,p_\theta(\mathbf{z}_i)\big),
\end{aligned}
\label{eq_mmvda}
\end{equation}
}
\noindent
\textcolor{black}{where $\beta$ controls the strength of the prior regularization term $L_{prior}$ as in the $\beta$-VAE~\cite{Higgins2017Beta-VAE:Framework} and $\gamma=10$ is a fixed scaling parameter. As in VDA~\cite{Ziogas2026VariationalRepresentations}, $\mathbf{\tilde{z}}_0$ is not regularized by $\mathcal{L}_{prior}$, and the posterior means form the utilized representation.}

\textcolor{black}{\textbf{Component-wise conditional decoder.} MMVDA extends VDA by also learning the mappings $g_i$ of Eq.~\eqref{eq_gen_model_dists} and enables the generation of multi-modal observations. As the noise is independent across components and $q_\phi$ in Eq.~\eqref{eq_posterior_model} factorizes exactly over the components, the evidence lower bound (ELBO) for the auto-encoder instantiated by the MMVDA variant with a decoder, splits into per-component terms:}

\textcolor{black}{
\begin{equation}\label{eq_elbo}
\begin{aligned}
\ln p_\theta(\mathbf{x})\ge\textstyle\sum_{i=1}^{C}\Big(&\mathbb{E}_{q_\phi(\mathbf{z}_i\mid\mathbf{x}^{(i)})}\big[\ln p_\theta(\mathbf{x}^{(i)}\mid\mathbf{z}_i)\big]\\
&-D_{KL}\big(q_\phi(\mathbf{z}_i\mid\mathbf{x}^{(i)})\,\|\,p_\theta(\mathbf{z}_i)\big)\Big).
\end{aligned}
\end{equation}
}
\noindent
\textcolor{black}{where notably, a single shared decoder suffices to learn mappings $g_i$ by providing a learned embedding $\mathbf{c}_i$ that conditions $\mathbf{z}_i$ with the identity of component $i$ and hence $g_i(\mathbf{z}_i)=g_\theta(\mathbf{z}_i,\mathbf{c}_i)$ (see Section~\ref{decoder_theory}). Under Eqs.~\eqref{eq_joint_probability}-\eqref{eq_posterior_model} and Eq.~\ref{eq_elbo}, each component is decoded from its own latent only and generated components occupy their own subspaces, whereas the anchor latent is not decoded. The MMVDA objective with a decoder is given as:}

\textcolor{black}{
\begin{equation}
\begin{aligned}
\mathcal{L}^{dec}_{MMVDA}&=\lambda_{dec}\,\mathcal{L}_{dec}+\mathcal{L}_{recon}+\mathcal{L}_{ortho}+\beta\,\mathcal{L}_{prior},\\
\mathcal{L}_{dec}=-\tfrac{1}{B}\textstyle&\sum_{n,i}\mathbb{E}_{q_\phi(\mathbf{z}_i\mid\mathbf{x}^{(i)}_n)}\big[\ln p_\theta\big(\mathbf{x}^{(i)}_n\mid\mathbf{z}_i,\mathbf{c}_i\big)\big].
\end{aligned}
\label{eq_mmvda_dec}
\end{equation}
}
\noindent
where $\lambda_{dec}=0.5$ weights the negative ELBO term of Eq.~\eqref{eq_elbo} implemented as the smooth-$L_1$ loss of Section~\ref{decoder_theory}, and $\mathcal{L}_{recon},\mathcal{L}_{ortho},\mathcal{L}_{prior}$ enforce cross-modal alignment, orthogonality in $\mathcal{H}$, and regularization of the component subspaces. Because the prior $p_\theta$ factorizes over components, components are encoded, sampled and generated independently, and we evaluate generation per component and report the sum. We denote the models trained with Eqs.~\eqref{eq_mmvda},\eqref{eq_mmvda_dec}, as $OmniDecVAE_{Enc}$ and $OmniDecVAE_{Enc-Dec}$.

\begin{figure*}[ht!]
    \includegraphics[height = 8.5cm,width=0.85\textwidth]{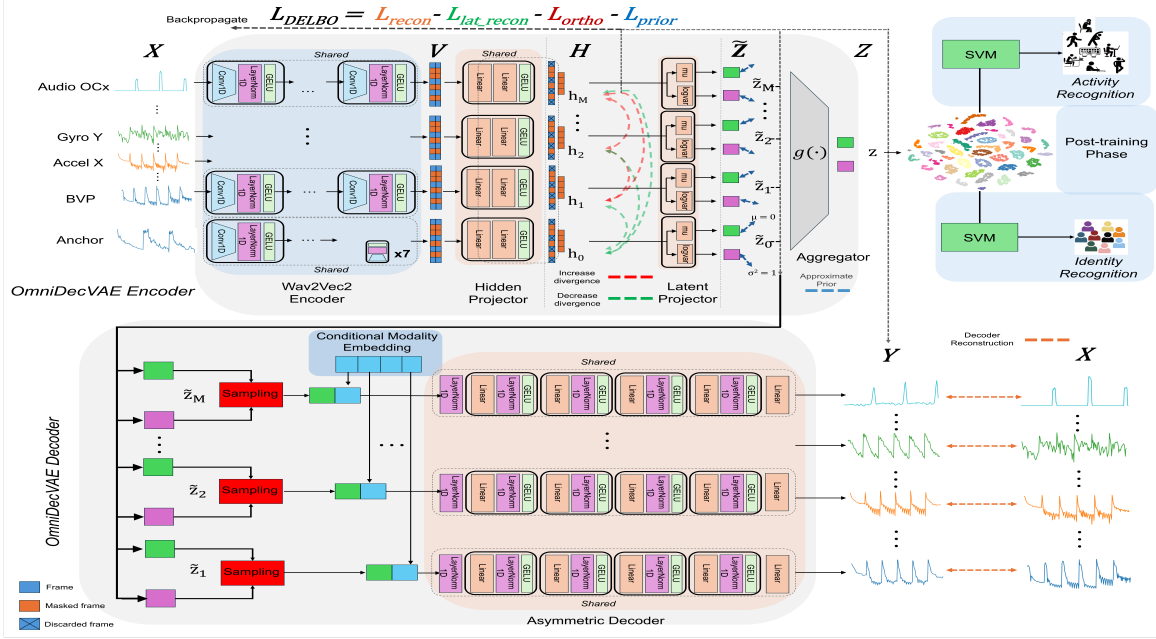}
    \centering
    \caption{Neural network architecture of the proposed OmniDecVAE scheme, that applies MMVDA.}
    \label{fig:architecture}
\end{figure*}

\subsection{OmniDecVAE: Omni-modal Variational Decomposition Autoencoder}
\label{omnidecvae_methods}
Fig.~\ref{fig:architecture} illustrates the OmniDecVAE architecture. The TF inputs of all components and the anchor are processed by a shared encoder $f_W$, a five-layer one-dimensional CNN based on Wav2vec2~\cite{Baevski2020Wav2vecRepresentations} with LayerNorm~\cite{Ba2016LayerNormalization}, GELU activations~\cite{Hendrycks2016GaussianGELUs}, and a shared FC projection head $f_H$ maps them to the embedding space $\mathcal{H}$, where $\mathcal{L}_{SSLDec}$ (Eq.~\eqref{eq_decomp_loss_compute}) is applied. $C+1$ FC projection layers then output the means and log-variances of the Gaussian component subspaces $\mathcal{Z}_i$, and $\mathcal{L}_{prior}$ is computed for every component subspace but not for the anchor. The subspaces are concatenated into the representation $\mathbf{\tilde{z}}_0 = [\mathbf{\tilde{z}}_0;\mathbf{\tilde{z}}_1;\dots;\mathbf{\tilde{z}}_C]$ of dimension $z_{dim} \times (C+1)$. The design of OmniDecVAE thus handles multi-modality entirely by the objective; there are no modality-specific encoders, and modality-specific structure is enforced only at the latent projectors. The hyperparameter $l_{SSLDec\%}=50$ sets the percentage of latent frames used in $\mathcal{L}_{SSLDec}$ via random masking, acting as a regularizer.

\subsection{Asymmetric Modality-Conditioned Shared Decoder}
\label{decoder_theory}

The generative model $\theta$ is a four-layer FC decoder with LayerNorm~\cite{Ba2016LayerNormalization}, GELU activations~\cite{Hendrycks2016GaussianGELUs} and a linear output layer, which makes OmniDecVAE an asymmetric autoencoder: as in masked autoencoders~\cite{he2022MAE}, the encoder remains expressive while the decoder is small. The anchor latent $\mathbf{\tilde{z}}_0$ is not decoded, so reconstruction relies solely on the component latents (Section~\ref{mmvda_objective}). To sample from $q_\phi(\mathbf{z}_i \mid \mathbf{x}^{(i)})$, we employ the reparameterization trick \cite{Kingma2014Auto-EncodingBayes}:
\begin{equation}
\tilde{\mathbf{z}}_i=\boldsymbol{\mu}_{z_i}+\tau\,\boldsymbol{\sigma}_{z_i}\odot\boldsymbol{\epsilon},\qquad \boldsymbol{\epsilon}\sim\mathcal{N}(\mathbf{0},\mathbf{I}),\quad i\in\{1,\dots,C\},
\label{eq_reparam_trick}
\end{equation}
\noindent
where $\tau$ scales the stochastic term: $\tau=1$ gives the standard reparameterization, whereas $\tau<1$ acts as a temperature on the posterior, i.e., $\mathcal{N}(\mathbf{\mu}_{z_i},\tau^2diag\mathbf{\sigma^2_{z_i}})$. Empirically, smaller values of $\tau$ lead to more stable training as the number of modalities increased ($\tau=0.15$ for HARWE, $0.25$ for WESAD, and $0.5$ for UCI HAR). A learned conditional embedding $\mathbf{c}_i$ identifies component $i$ and informs the decoder of which component to reconstruct: the decoder input is $LayerNorm([\tilde{\mathbf{z}}_i, \mathbf{c}_i])$, and its output is the flattened TF representation of the component. For a component $x$ and its reconstruction $y$, $-\ln p_\theta$ in $\mathcal{L}_{dec}$ (Eq.~eq\ref{eq_mmvda_dec}) is the smooth-L1 loss summed over the TF features, i.e., $l_{dec}=0.5(x-y)^2 \ \text{if} \ |x_n-y_n| < 1, \ \text{and} \ |x_n-y_n| - 0.5 \ \text{otherwise}$.


\section{Experiments} \label{experiments}
\subsection{Datasets}
We utilize HARWE~\cite{ Esmaeilzehi2024HARWE:Environments}, a multi-modal HAR dataset that contains video, audio, inertial and physiological recording of 35 participants performing nine daily activities in work environments, with about 10 min per activity. We use all time-series modalities, for a total of $C=30$ components: monaural audio ($16kHz$), separated into six frequency-band OCs by filter decomposition~\cite{Ziogas2026VariationalRepresentations}; BVP ($64Hz$), EDA ($4Hz$), and Temperature (TEMP) ($64Hz$) from a smartwatch, and seven tri-axial inertial sensors ($200Hz$) from a smartphone, i.e., ACC, GYR, Gravity (GRA), Orientation (OR), Rotation (RO) Magnetometer (MG), Linear accelerometer (LACC).

To assess generalization beyond HARWE, we additionally analyze two multi-modal wearable datasets with a different set of sensors and tasks, UCI HAR~\cite{anguita2013public}, and WESAD~\cite{schmidt2018introducing}. UCI HAR~\cite{anguita2013public} contains tri-axial inertial signals (ACC, GYR, GRA, $C=9$) from a waist-worn smartphone, for 30 subjects performing six activities. WESAD~\cite{schmidt2018introducing} contains chest (RespiBAN: tri-axial ACC, ECG, EDA, EMG, respiration (RESP), TEMP) and wrist (tri-axial ACC, BVP, EDA, TEMP) recordings ($C=14$) of 15 subjects in the affective states of baseline, stress, amusement, and meditation. Pre-training in WESAD uses data from all four affective states, and evaluation is performed using labels from the three affective states without meditation.

On HARWE, the subject-dependent (SD; Easy in~\cite{Esmaeilzehi2024HARWE:Environments}) scenario uses $80\%$ of each subject's recording for training and $20\%$ for testing, and the subject-independent (SI; Difficult in~\cite{Esmaeilzehi2024HARWE:Environments}) scenario uses $70\%$ of the subjects for training and $30\%$ for testing. On UCI HAR the SI scenario uses the default 21/9 subject split~\cite{anguita2013public}, and for WESAD a fixed 10/5 subject split (test subjects S3, S8, S10, S11, S16); the SD scenario uses $20\%$ of each subject's recording as a test set.

\subsection{Comparisons and Benchmarks}
We compare the OmniDecVAE framework to a number of different approaches used on HARWE \cite{Esmaeilzehi2024HARWE:Environments}, standard benchmark approaches in the multi-modal fusion domain \cite{yamnet2019}, \cite{Lawhern2018EEGNet:Interfaces}, AE-based models \cite{Kingma2014Auto-EncodingBayes}, \cite{Higgins2017Beta-VAE:Framework}, \cite{Ziogas2026VariationalRepresentations} and eigenprojection models ICA \cite{hyvarinen2001independent}, PCA \cite{Greenacre2022PrincipalAnalysis}, and kernel-PCA variants \cite{Scholkopf1997KernelAnalysis} \textcolor{black}{with a kernel width of $\gamma= 1/(pVar(\mathbf{X}))$, with $p$ the number of input features; we report the sigmoid kernel, the most accurate kernel on HARWE}. We also instantiate a multi-modal VAE model (MMVAE), by using the same architecture as in Fig.\ref{fig:architecture}, but with modality-dedicated branches instead of a shared branch. 

\subsection{Implementation and Optimization}

We pre-process HARWE as in~\cite{Esmaeilzehi2024HARWE:Environments} and compute the STFT of all physiological and inertial channels in non-overlapping $3$ s frames (512-point STFT, $0.75$ s Hamming window, $75\%$ overlap), giving a $64\times3$ spectrogram flattened to 192 features; audio OCs use a Mel spectrogram with 64 Mel bands and 3 hops. UCI HAR is resampled to $75Hz$ with its default $2.56$ s frames ($50\%$ overlap), and WESAD to $64Hz$, except EMG ($700Hz$), in $3$ s frames. Every component, including the anchor, is represented as a 192-point STFT magnitude spectrogram. The anchor (Eq.~\eqref{eq_anchor}) is the energy preserving superposition of all standardized channels before the TF transform:

\begin{equation}
A(\mathbf{x})=\textstyle\sum_{c=1}^{C'}\mathcal{R}\big(\bar{\mathbf{x}}_{c}\big),
\label{eq_anchorsum}
\end{equation}

where $\bar{\mathbf{x}}_c$ is the standardized channel $c$, $\mathcal{R}(\cdot)$ resamples to the highest common rate, and $C'$ is the number of channels before audio is split into OCs. When acoustic and other modalities are mixed, the anchor uses the Mel transform instead, as a "sonified" representation of all channels. All signals are normalized using batch statistics.

\textcolor{black}{For the weak contrastive pairs of Eq.~\eqref{eq_asymm_weights}, we follow the rules laid out in Section~\ref{omni_modal_contrasting} for HARWE. On UCI HAR, the three axes of the same sensor and the corresponding axes of different sensors are weak contrastive pairs. On WESAD they are the axes of the two ACC (chest, wrist) signals, same device physiology (chest ECG, EDA, EMG, RESP, TEMP; wrist BVP, EDA, TEMP) and same physiology but different device. Because neither UCI HAR or WESAD have audio, the positive weights in Eq.~\ref{eq_asymm_weights} are kept uniform ($IF_{pos}=0$).}

\textcolor{black}{OmniDecVAE uses the same architecture and hyperparameters on the three datasets. The encoder has five convolutional layers of $512$ channels (kernel sizes of $[8,3,3,2,2]$, strides of $[4,4,3,2,2]$), and the projection head maps its 512-dimensional output to $\mathcal{H}$ ($d=192$); each latent subspace $\mathcal{Z}_i$ has 32 dimensions. The decoder has four FC layers ($[128,256,512,512]$ units) and an 8-dimensional conditional embedding $\mathbf{c}_i$. In the experiments with HARWE components (Section~\ref{modalities_ablation}), $\tau$ of Eq.~\eqref{eq_reparam_trick} is selected as $[1,0.8,0.5,0.1,0.1,0.25,0.15]$ for $C=[3,6,9,21,24,27,30]$, respectively.} We optimize our networks on a single NVIDIA RTX 6000 Ada Generation using PyTorch~\cite{paszke2019pytorch} and Huggingface~\cite{wolf2020HFtransformers} using the Adam optimizer~\cite{kingma2014Adam} \textcolor{black}{($b_1=0.5,b_2=0.999,\epsilon=10^{-6}$, L2 penalty $10^{-4}$), for 120 epochs, with a batch size of $64$ and we clip gradients with a norm larger than 1 (OmniDecVAE$_{Enc-Dec}$) or 2 (OmniDecVAE$_{Enc}$); a linear warm-up schedule with peak learning rate of $8\cdot10^{-5}$ (OmniDecVAE$_{Enc}$), $10^{-5}$ (OmniDecVAE$_{Enc-Dec}$), followed by linear decay, is used}. Other VAE-based models are trained for 100 epochs as their optimization converges faster. Our learning scheme consists of a pre-training stage, where the model is trained to optimize one of Eqs.~(\eqref{eq_mmvda},\eqref{eq_mmvda_dec}), and a post-training stage where a simple Support Vector Machine (SVM) classifier~\cite{pedregosa2011scikit}) is utilized as a classification oracle on the learned representations. For combined CNN-Transformer-based fusion, we use the training schemes for supervised learning from Esmailzehi \textit{et al.}~\cite{Esmaeilzehi2024HARWE:Environments}.

\subsection{Evaluation metrics}
\subsubsection{Representation informativeness}
HARWE allows us to effectively perform the supervised classification tasks of HAR (on HARWE and UCI HAR), identity recognition (IR) (on HARWE, UCI HAR, and WESAD), and affect recognition (AR: baseline, stress, amusement) (on WESAD). We evaluate these tasks using Accuracy and the F1-score (F1). For the SD scenario we evaluate for HAR,  IR, and AR, whereas in the SI scenario we evaluate only for HAR or AR. 

\subsubsection{Disentanglement quality}
\textcolor{black}{We measure disentanglement using unsupervised and supervised metrics that utilize the labels of each dataset as the generative factors. Modality is not treated as a factor as each latent subspace is tied to a known component by construction. In this direction, we report four metrics: Normalized Gaussian correlation (GCN)~\cite{Locatello2019ChallengingRepresentations} measures the statistical dependence between latent dimensions. Moreover, supervised disentanglement using the DCI framework's~\cite{Eastwood2018ARepresentations} disentanglement (D), and informativeness (I)), and the interventional robustness score (IRS)~\cite{suter2019irs}. For DCI, random forests are used to obtain importances and calculate D and I. All metrics are computed on the test sets with a stratified 5-fold cross-validation on the full representations of all models (including the anchor in OmniDecVAEs).}

\subsubsection{Generation quality}
We sample from the latent noise distribution of the learned conditional subspaces to generate a set of multi-modal signals $Y_{gen}$, and compare it with a subset $X_{real}$ of the test set through the Multi-Kernel-Maximum Mean Discrepancy (MK-MMD), which is the MMD averaged over RBF kernels of multiple bandwidths~\cite{gretton2012kernel}. We also compute the diversity score $DivScore = | d_{real} - d_{gen}|$, which is the difference between the mean pairwise Euclidean distances within the two sets. For reconstruction, we compare the TF inputs $X_n$ with their reconstructions $Y_n$ through the mean squared and absolute errors (MSE, MAE), and the latents $Z_n$ with the latents $Z_{Y_n}$ of the re-encoded reconstructions, through the $L2$ and cosine distances ($d_{L2}, d_{cos}$) which measure latent consistency.

\subsubsection{Utility of generated data}
Because the decoders of OmniDecVAEs and the MMVAE branches are conditioned on modality and not task labels, we obtain task-labeled synthetic windows by fitting a Gaussian with diagonal covariance to the posterior means of every component subspace of the labeled training frames, sampling from it and decoding every component. A two-layer MLP is trained on $10\%,50\%$, or $100\%$ of the real SI training windows, alone or augmented with windows generated by $\beta$-OmniDecVAE$_{Enc-Dec}$, $\beta$-MMVAE, or jittering and scaling, or by training on synthetic frames only. Testing is done on the real SI test set.

\subsubsection{Complexity}
To evaluate processing and deployment relevant properties, we measure floating-point operations (FLOPs), number of parameters (Param.), size on disk (SoD), and frames processed per second during inference (latency).

\subsubsection{Statistical analysis}
All results are mean $\pm95\%$ confidence interval (CI) ($t$-distribution) over independent runs: 5 pretraining seeds $\times$ 5 classifier states for the representation models, and 25 seeds for the end-to-end models in recognition, and 5 pretraining seeds (25 for end-to-end models) in disentanglement and generation. The generated data utility experiment gets 5 $\times$ 5 generator and classifier seeds. Results tables mark bold as the best model, and underline as the models not significantly different from it (Welch $t$-test, $\alpha=0.05$).

\section{Results and Discussion}
\label{results}

\begin{table*}[!t]\centering
\caption{\textcolor{black}{Omni-modal recognition accuracy (\%) on HARWE~\cite{Esmaeilzehi2024HARWE:Environments}, UCI HAR~\cite{anguita2013public}, and WESAD~\cite{schmidt2018introducing}}}
\label{tab:recognition}
\scriptsize\setlength{\tabcolsep}{3.0pt}\begin{adjustbox}{max width=\textwidth}%
{\color{black}
\begin{tabular}{lccccccccc}\toprule
 & \multicolumn{3}{c}{HARWE~\cite{Esmaeilzehi2024HARWE:Environments} ($C=30$)} & \multicolumn{3}{c}{UCI HAR~\cite{anguita2013public} ($C=9$)} & \multicolumn{3}{c}{WESAD~\cite{schmidt2018introducing} ($C=14$)}\\
\cmidrule(lr){2-4}\cmidrule(lr){5-7}\cmidrule(lr){8-10}
 & \multicolumn{2}{c}{SD} & SI & \multicolumn{2}{c}{SD} & SI & \multicolumn{2}{c}{SD} & SI\\
\cmidrule(lr){2-3}\cmidrule(lr){4-4}\cmidrule(lr){5-6}\cmidrule(lr){7-7}\cmidrule(lr){8-9}\cmidrule(lr){10-10}
Model & HAR & IR & HAR & HAR & IR & HAR & AR & IR & AR\\\midrule
Random Classifier & $11.2 \pm 0.2$ & $2.9 \pm 0.1$ & $11.1 \pm 0.2$ & $17.0 \pm 1.1$ & $3.6 \pm 0.5$ & $15.8 \pm 0.8$ & $32.9 \pm 2.4$ & $6.6 \pm 0.7$ & $33.5 \pm 0.5$\\
Majority Classifier & $11.5 \pm 0.0$ & $3.4 \pm 0.0$ & $11.2 \pm 0.0$ & $19.0 \pm 0.0$ & $3.9 \pm 0.0$ & $18.2 \pm 0.0$ & $53.3 \pm 0.0$ & $6.9 \pm 0.0$ & $52.8 \pm 0.0$\\
\midrule
Logistic Regression & $38.1 \pm 0.1$ & $18.5 \pm 0.4$ & $38.3 \pm 0.1$ & $78.0 \pm 0.1$ & $12.5 \pm 1.5$ & $72.6 \pm 0.6$ & $68.2 \pm 0.1$ & $36.6 \pm 2.1$ & $64.2 \pm 0.2$\\
Random Forest & $44.5 \pm 0.2$ & $25.2 \pm 0.1$ & $43.8 \pm 0.1$ & $76.0 \pm 0.6$ & $25.0 \pm 0.7$ & $73.5 \pm 0.6$ & $69.7 \pm 0.4$ & $35.5 \pm 0.8$ & $63.9 \pm 0.3$\\
SVM & $42.6 \pm 0.1$ & $17.3 \pm 0.1$ & $43.1 \pm 0.1$ & $79.1 \pm 0.3$ & $19.2 \pm 0.4$ & $75.3 \pm 0.4$ & $69.7 \pm 0.3$ & $36.5 \pm 0.3$ & $63.5 \pm 0.1$\\
\midrule
ICA~\cite{hyvarinen2001independent} & $77.0 \pm 0.1$ & $75.3 \pm 0.1$ & $62.1 \pm 0.1$ & $88.1 \pm 0.1$ & $29.3 \pm 0.2$ & $85.2 \pm 0.1$ & $93.7 \pm 0.1$ & $93.8 \pm 0.1$ & $72.8 \pm 0.1$\\
PCA~\cite{Greenacre2022PrincipalAnalysis} & $77.0 \pm 0.1$ & $75.3 \pm 0.1$ & $62.1 \pm 0.1$ & $88.0 \pm 0.1$ & $29.2 \pm 0.3$ & $85.3 \pm 0.1$ & $93.7 \pm 0.1$ & $93.8 \pm 0.1$ & $72.8 \pm 0.1$\\
sigmoid-PCA~\cite{Scholkopf1997KernelAnalysis} & $77.3 \pm 0.1$ & $74.9 \pm 0.1$ & $61.4 \pm 0.1$ & $86.9 \pm 0.2$ & $29.3 \pm 0.2$ & $84.1 \pm 0.1$ & $92.9 \pm 0.1$ & $93.2 \pm 0.1$ & $\mathbf{75.1 \pm 0.1}$\\
\midrule
CNN LF, Single-Head & $38.1 \pm 5.7$ & $50.2 \pm 4.6$ & $31.9 \pm 4.0$ & $\mathbf{95.7 \pm 0.4}$ & $\mathbf{61.5 \pm 0.5}$ & $\mathbf{93.1 \pm 0.4}$ & $95.7 \pm 1.6$ & $\underline{98.4 \pm 0.6}$ & $46.9 \pm 2.1$ \\
CNN LF, Dual-Head & $49.9 \pm 4.9$ & $50.1 \pm 4.5$ & NA & $94.5 \pm 0.5$ & $\underline{61.2 \pm 0.7}$ & NA & $\mathbf{98.5 \pm 0.2}$ & $\mathbf{98.9 \pm 0.3}$ & NA\\
CNN-Transformer~\cite{Esmaeilzehi2024HARWE:Environments}, Single-Head & $39.9 \pm 6.3$ & $67.4 \pm 3.1$ & $33.0 \pm 2.4$ & $93.0 \pm 0.6$ & $58.1 \pm 0.8$ & $92.0 \pm 0.5$ & $80.7 \pm 6.2$ & $90.3 \pm 8.0$ & $52.3 \pm 2.8$\\
CNN-Transformer~\cite{Esmaeilzehi2024HARWE:Environments}, Dual-Head & $53.9 \pm 3.1$ & $61.6 \pm 3.7$ & NA & $92.2 \pm 0.5$ & $58.2 \pm 0.6$ & NA & $86.3 \pm 6.5$ & $82.4 \pm 9.8$ & NA\\
\midrule
MMVAE & $83.5 \pm 0.1$ & $83.2 \pm 0.2$ & $63.1 \pm 0.1$ & $81.5 \pm 0.2$ & $22.6 \pm 0.2$ & $79.0 \pm 0.2$ & $90.3 \pm 0.3$ & $91.7 \pm 0.3$ & $70.2 \pm 0.4$\\
$\beta$-MMVAE ($\beta = 0.1$) & $82.4 \pm 0.1$ & $79.4 \pm 0.4$ & $\mathbf{70.9 \pm 0.1}$ & $84.5 \pm 0.3$ & $28.5 \pm 0.2$ & $81.6 \pm 0.3$ & $90.4 \pm 0.2$ & $93.1 \pm 0.2$ & $72.7 \pm 0.4$\\
\midrule
OmniDecVAE$_{Enc}$~\cite{Ziogas2026VariationalRepresentations} & $71.5 \pm 1.2$ & $62.1 \pm 2.8$ & $61.9 \pm 0.4$ & $77.0 \pm 0.6$ & $19.6 \pm 0.9$ & $69.4 \pm 0.9$ & $89.1 \pm 0.5$ & $82.4 \pm 1.3$ & $63.4 \pm 0.9$\\
$\beta$-OmniDecVAE$_{Enc}$~\cite{Ziogas2026VariationalRepresentations} ($\beta = 80$) & $71.8 \pm 1.0$ & $63.1 \pm 1.3$ & $59.3 \pm 1.1$ & $75.7 \pm 1.7$ & $19.2 \pm 1.8$ & $70.4 \pm 1.3$ & $94.5 \pm 0.3$ & $91.9 \pm 0.8$ & $65.6 \pm 1.0$\\
OmniDecVAE$_{Enc-Dec}$ & $82.2 \pm 0.2$ & $86.1 \pm 0.1$ & $60.8 \pm 0.2$ & $81.8 \pm 0.7$ & $27.8 \pm 0.4$ & $76.9 \pm 0.5$ & $95.2 \pm 0.2$ & $93.9 \pm 0.2$ & $70.7 \pm 0.3$\\
$\beta$-OmniDecVAE$_{Enc-Dec}$ ($\beta = 0.1$) & $\mathbf{84.6 \pm 0.2}$ & $\mathbf{89.0 \pm 0.1}$ & $64.6 \pm 0.1$ & $84.1 \pm 0.3$ & $32.5 \pm 0.3$ & $79.8 \pm 0.3$ & $\mathbf{96.2 \pm 0.1}$ & $\mathbf{95.6 \pm 0.1}$ & $74.6 \pm 0.4$\\
\bottomrule\end{tabular}}\end{adjustbox}
\par\smallskip\raggedright\scriptsize \textcolor{black}{$n=25$ (5 pre-training seeds $\times$ 5 classifier states), supervised end-to-end models: 25 training seeds, mean $\pm95\%$ CIs reported. CNN LF: YAMNet~\cite{yamnet2019} \& EEGNet~\cite{Lawhern2018EEGNet:Interfaces} feature extractors with late fusion; CNN-Transformer without the YAMNet (audio) branch on UCI HAR and WESAD. Bold: best per column. Underlined: not significantly different from the best (Welch $t$-test, $p\geq0.05$).}
\end{table*}

\begin{table*}[!t]\centering
\caption{\textcolor{black}{Disentanglement evaluation on the SI split of HARWE~\cite{Esmaeilzehi2024HARWE:Environments}, UCI-HAR~\cite{anguita2013public} and WESAD~\cite{schmidt2018introducing}}}
\label{tab:disentanglement}
\scriptsize\setlength{\tabcolsep}{2.0pt}\begin{adjustbox}{max width=\textwidth}%
{\color{black}
\begin{tabular}{lcccccccccccc}\toprule
 & \multicolumn{4}{c}{HARWE ($C=30$)} & \multicolumn{4}{c}{UCI HAR ($C=9$)} & \multicolumn{4}{c}{WESAD ($C=14$)}\\
\cmidrule(lr){2-5}\cmidrule(lr){6-9}\cmidrule(lr){10-13}
Model & D $\uparrow$ & I $\uparrow$ & IRS $\uparrow$ & GCN $\downarrow$ & D $\uparrow$ & I $\uparrow$ & IRS $\uparrow$ & GCN $\downarrow$ & D $\uparrow$ & I $\uparrow$ & IRS $\uparrow$ & GCN $\downarrow$\\\midrule
PCA [38] & $4.4 \pm 0.0$ & $77.3 \pm 0.2$ & $39.6 \pm 0.8$ & $0.01 \pm 0.00^{\dagger}$ & $6.3 \pm 0.1$ & $72.5 \pm 0.6$ & $44.2 \pm 0.3$ & $0.00 \pm 0.00^{\dagger}$ & $6.2 \pm 0.1$ & $96.0 \pm 0.1$ & $43.7 \pm 0.3$ & $0.03 \pm 0.00^{\dagger}$\\
ICA [22] & $5.6 \pm 0.2$ & $86.7 \pm 0.7$ & $44.7 \pm 1.0$ & $0.89 \pm 0.03$ & $7.6 \pm 0.3$ & $70.6 \pm 0.3$ & $43.7 \pm 0.7$ & $2.03 \pm 0.35$ & $7.3 \pm 0.7$ & $97.9 \pm 0.1$ & $39.6 \pm 0.3$ & $3.64 \pm 0.07$\\
\midrule
CNN-Transformer [5] & $5.5 \pm 0.5$ & $\mathbf{97.7 \pm 0.4}$ & $36.9 \pm 1.0$ & $2.48 \pm 0.16$ & $12.2 \pm 0.7$ & $\mathbf{87.7 \pm 0.5}$ & $\underline{53.7 \pm 1.9}$ & $10.17 \pm 0.63$ & $9.9 \pm 2.0$ & $86.4 \pm 5.1$ & $35.3 \pm 3.8$ & $1.71 \pm 0.20$\\
\midrule
MMVAE & $29.3 \pm 0.6$ & $91.6 \pm 0.2$ & $50.3 \pm 0.5$ & $0.70 \pm 0.03$ & $11.1 \pm 1.2$ & $73.4 \pm 0.2$ & $\underline{48.4 \pm 0.8}$ & $0.56 \pm 0.03$ & $21.6 \pm 1.6$ & $98.3 \pm 0.2$ & $47.4 \pm 0.3$ & $0.52 \pm 0.02$\\
$\beta$-MMVAE & $23.3 \pm 0.7$ & $90.9 \pm 0.4$ & $47.2 \pm 0.7$ & $0.60 \pm 0.01$ & $\underline{12.8 \pm 1.4}$ & $71.3 \pm 0.3$ & $38.7 \pm 0.6$ & $0.80 \pm 0.04$ & $21.2 \pm 1.5$ & $\mathbf{98.6 \pm 0.1}$ & $42.7 \pm 0.7$ & $0.57 \pm 0.01$\\
\midrule
OmniDecVAE$_{Enc}$ & $14.4 \pm 1.7$ & $70.0 \pm 2.6$ & $\mathbf{55.5 \pm 2.4}$ & $\mathbf{0.03 \pm 0.00}$ & $11.8 \pm 1.9$ & $50.5 \pm 1.8$ & $\mathbf{57.6 \pm 12.6}$ & $\mathbf{0.05 \pm 0.02}$ & $14.6 \pm 1.9$ & $92.1 \pm 1.2$ & $\mathbf{90.9 \pm 2.7}$ & $\mathbf{0.05 \pm 0.02}$\\
$\beta$-OmniDecVAE$_{Enc}$ & $15.5 \pm 3.4$ & $69.3 \pm 5.8$ & $\underline{55.3 \pm 18.1}$ & $\underline{0.10 \pm 0.09}$ & $11.0 \pm 2.0$ & $55.3 \pm 3.6$ & $\underline{44.7 \pm 9.2}$ & $\underline{0.09 \pm 0.05}$ & $15.3 \pm 2.0$ & $93.9 \pm 3.2$ & $\underline{90.5 \pm 8.7}$ & $0.10 \pm 0.04$\\
OmniDecVAE$_{Enc-Dec}$ & $28.2 \pm 0.6$ & $87.1 \pm 0.3$ & $39.1 \pm 0.8$ & $0.46 \pm 0.01$ & $12.6 \pm 0.4$ & $69.5 \pm 0.5$ & $\underline{53.5 \pm 2.5}$ & $0.40 \pm 0.08$ & $22.9 \pm 1.0$ & $97.5 \pm 0.1$ & $43.2 \pm 0.9$ & $0.35 \pm 0.01$\\
$\beta$-OmniDecVAE$_{Enc-Dec}$ & $\mathbf{30.4 \pm 0.6}$ & $88.9 \pm 0.3$ & $39.4 \pm 0.8$ & $0.66 \pm 0.01$ & $\mathbf{13.8 \pm 1.1}$ & $73.6 \pm 0.4$ & $\underline{49.7 \pm 1.2}$ & $0.60 \pm 0.11$ & $\mathbf{24.6 \pm 1.0}$ & $98.5 \pm 0.1$ & $41.2 \pm 1.6$ & $0.54 \pm 0.01$\\
\bottomrule\end{tabular}}\end{adjustbox}
\par\smallskip\raggedright\scriptsize \textcolor{black}{D, I: DCI~\cite{Eastwood2018ARepresentations} disentanglement and informativeness; IRS: interventional robustness~\cite{suter2019irs} (all $\times100$); GCN: normalized Gaussian correlation of the latent dimensions~\cite{Locatello2019ChallengingRepresentations}. Factors: activity (affect for WESAD) and identity. $n=5$ pre-training seeds ($n=25$ for the CNN-Transformer). Mean $\pm$ 95\% CI ($t$-distribution) is reported. Bold: best per column. Underlined: not significantly different from the best (Welch $t$-test, $p\geq0.05$). $^{\dagger}$PCA decorrelates its dimensions by construction (GCN not compared).}
\end{table*}

\subsection{Representation Informativeness and Disentanglement}
Table \ref{tab:recognition} tabulates recognition performance when using all modalities. On HARWE, $\beta$-OmniDecVAE$_{Enc-Dec}$ is the most accurate model in the SD scenario for both HAR and IR ($84.6\%$ and $89.0\%$; VAE baselines~\cite{Higgins2017Beta-VAE:Framework,Kingma2014Auto-EncodingBayes} up to $83.5\%$ and $83.2\%$), ahead of the eigen-projections~\cite{hyvarinen2001independent,Greenacre2022PrincipalAnalysis,Scholkopf1997KernelAnalysis} and of the supervised CNN and CNN-Transformer~\cite{yamnet2019,Lawhern2018EEGNet:Interfaces,Esmaeilzehi2024HARWE:Environments}, which are also far more sensitive to the random seed despite their end-to-end training. In the SI scenario, $\beta$-MMVAE is the most accurate model ($70.9\%$ against $64.6\%$), which suggests that its multi-branch architecture adapts better to the unseen subjects of HARWE. On WESAD, the supervised late-fusion CNN is the most accurate model in the SD scenario (dual-head: $98.5\%$ AR and $98.9\%$ IR), and $\beta$-OmniDecVAE$_{Enc-Dec}$ is more accurate than $\beta$-MMVAE ($96.2\%$ and $95.6\%$, vs $90.4\%$ and $93.1\%$). In SI AR, $\beta$-OmniDecVAE$_{Enc-Dec}$ ($74.6\%$) is ahead of the supervised CNN and CNN-Transformer ($46.9\%$ and $52.3\%$), and of $\beta$-MMVAE ($72.7\%$), and second to sigmoid-PCA ($75.1\%$). On UCI HAR, the supervised models are the strongest (CNN LF: $95.7\%$ SD and $93.1\%$ SI HAR; CNN-Transformer: $93.0\%$ and $92.0\%$) followed by the eigen-projections ($86.9\%$ to $88.1\%$ SD, $84.1\%$ to $85.3\%$ SI). Among the VAE-based models, $\beta$-OmniDecVAE$_{Enc-Dec}$ and $\beta$-MMVAE are close in HAR ($84.1\%$ vs $84.5\%$ SD, $79.8\%$ vs $81.6\%$ SI) and $\beta$-OmniDecVAE$_{Enc-Dec}$ is better in IR ($32.5\%$ vs $28.5\%$). Supervised end-to-end training is thus the most accurate when subjects are known on UCI HAR and WESAD, but not on HARWE; for unseen subjects, the self-supervised models are favored on the heterogeneous HARWE ($C=30$) and WESAD ($C=14$), whereas the supervised models are favored on the single-device UCI HAR. On all three datasets, the decoder generally improves recognition in OmniDecVAEs.

Table~\ref{tab:disentanglement} presents disentanglement performance in the SI split of the three datasets. $\beta$-OmniDecVAE scores highest in DCI D in all three datasets ($D=30.4, 13.8$, and $24.6$ in HARWE, UCI HAR and WESAD; best baselines: MMVAE with $29.3$ and $21.6$ on HARWE and WESAD, $\beta$-MMVAE with $12.8$ on UCI HAR), while DCI informativeness stays within $2.7$ points of the VAE baselines ($I=88.9,73.6,98.5$ in the three datasets). The encoder-only OmniDecVAEs have the least dependent latent dimensions among the deep models (GCN$\leq0.10$) and the highest IRS ((OmniDecVAE$_{Enc}$ IRS $55.5$, $57.6$, $90.9$), but the lowest informativeness ($I\leq70.0,55.3,\ \text{and} \ 93.9$).

\begin{figure*}[ht!]
    \includegraphics[height=8cm,width=0.9\textwidth]{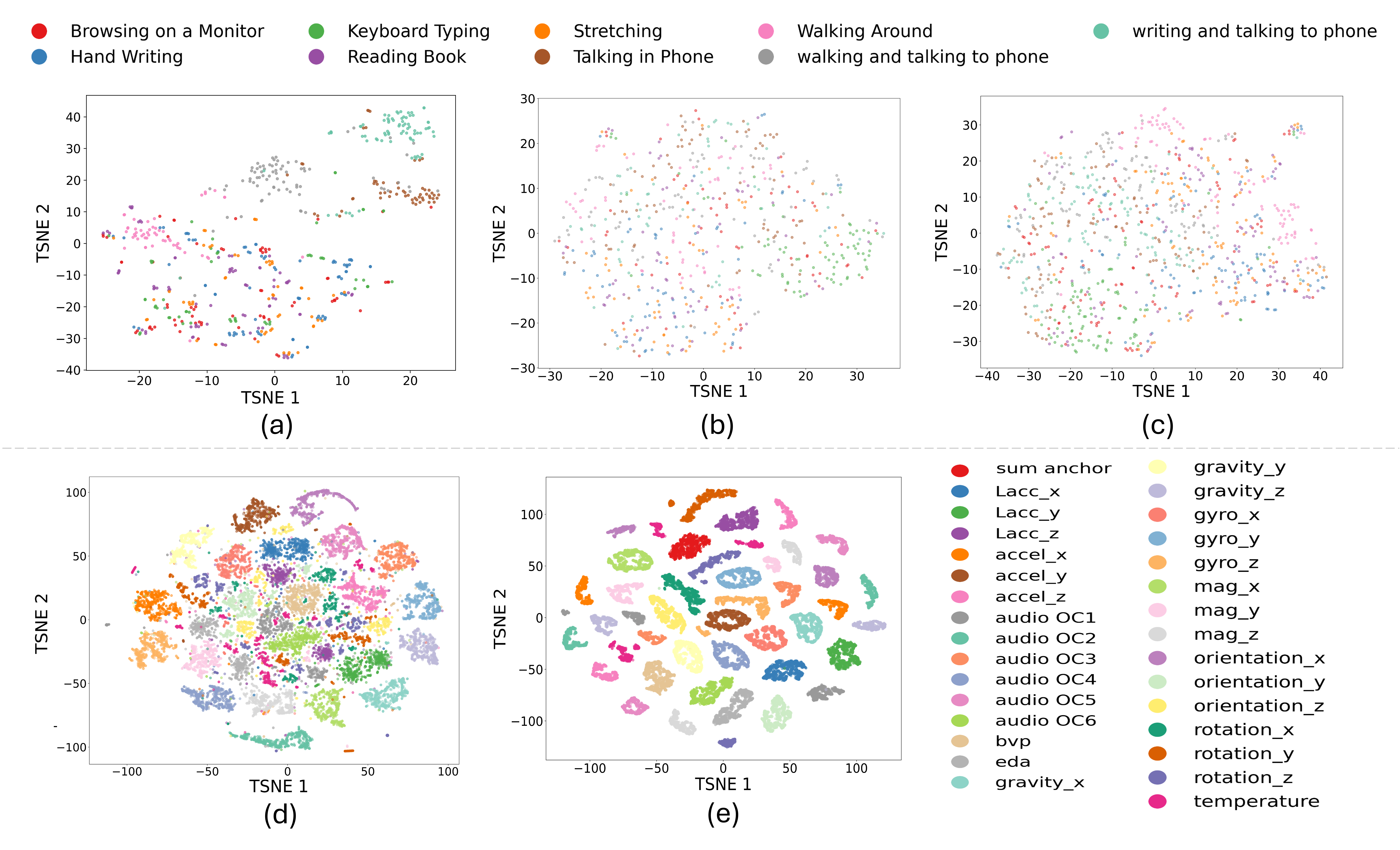}
    \centering
    \caption{TSNE \cite{maaten2008visualizing} visualizations of the learned latent representations \textcolor{black}{on HARWE~\cite{Esmaeilzehi2024HARWE:Environments}}. (a) Activity-colored latent from the CNN-Transformer \cite{Esmaeilzehi2024HARWE:Environments}, taken at the output of the Transformer layers. (b) Activity-colored latent from $\beta$-MMVAE. (c) Activity-colored latent from $\beta$-OmniDecVAE$_{Enc-Dec}$. (d) Modality-colored latent from $\beta$-MMVAE. (e) Modality-colored latent from $\beta$-OmniDecVAE$_{Enc-Dec}$. }
    \label{fig:tsne} 
\end{figure*}

Fig. \ref{fig:tsne} illustrates the \textcolor{black}{latent representations} of transformer-based \cite{Esmaeilzehi2024HARWE:Environments}, VAE-based \cite{Kingma2014Auto-EncodingBayes} and OmniDecVAE models \textcolor{black}{on HARWE}. It can be seen that the CNN-Transformer model achieves partial linear separation of activities when trained end-to-end for that task (Fig.~\ref{fig:tsne}a), whereas $\beta$-MMVAE and $\beta$-OmniDecVAE ((Fig. \ref{fig:tsne}b,c) learn unsupervised representations that are not linearly separable. However, $\beta$-MMVAE and $\beta$-OmniDecVAE achieve disentanglement in the sense of a structured modality space, as evident by Figs.\ref{fig:tsne}d,e. Notably, $\beta$-OmniDecVAE achieves \textcolor{black}{a clearer} separation between different modality modes (Fig.\ref{fig:tsne}e) in contrast to $\beta$-MMVAE (Fig. \ref{fig:tsne}d), which we attribute to the component-wise MMVDA objective (Eq.~\eqref{eq_mmvda_dec}). Each modality's location is \textcolor{black}{set} by the SSLDec structural dynamics of Eq.~\eqref{eq_decomp_loss_compute}, \textcolor{black}{through its alignment} to the anchor (latent reconstruction) and \textcolor{black}{its separation from} the other modalities (orthogonality). We see that the asymmetric weights \textcolor{black}{of Eq.~\eqref{eq_asymm_weights}} move the anchor (`center of mass' of the representation) away from some modalities and closer to others. 

\subsection{Generative Performance}

\begin{table*}[!t]\centering
\caption{\textcolor{black}{Omni-modal generative performance, SI split on HARWE~\cite{Esmaeilzehi2024HARWE:Environments}, UCI HAR~\cite{anguita2013public}, and WESAD~\cite{schmidt2018introducing}}}
\label{tab:generation_quality}
\scriptsize\setlength{\tabcolsep}{3.5pt}\begin{adjustbox}{max width=\textwidth}%
{\color{black}
\begin{tabular}{lcccccc}\toprule
 & \multicolumn{4}{c}{Reconstruction quality} & \multicolumn{2}{c}{Latent noise generation}\\\cmidrule(lr){2-5}\cmidrule(lr){6-7}
Model & MSE $\downarrow$ & MAE $\downarrow$ & $d_{L2}$ $\downarrow$ & $d_{\cos}$ $\downarrow$ & MK-MMD $\downarrow$ & DivScore $\downarrow$\\
\midrule\multicolumn{7}{l}{\textit{HARWE ($C=30$)}}\\
MMVAE & $0.216 \pm 0.000$ & $0.236 \pm 0.000$ & $0.410 \pm 0.003$ & $0.115 \pm 0.004$ & $0.059 \pm 0.000$ & $\mathbf{2.416 \pm 0.055}$\\
$\beta$-MMVAE & $0.205 \pm 0.000$ & $0.208 \pm 0.000$ & $0.931 \pm 0.009$ & $0.084 \pm 0.000$ & $\mathbf{0.054 \pm 0.001}$ & $2.756 \pm 0.022$\\
OmniDecVAE & $0.083 \pm 0.001$ & $0.194 \pm 0.001$ & $\mathbf{0.170 \pm 0.001}$ & $0.161 \pm 0.006$ & $0.100 \pm 0.005$ & $3.691 \pm 0.701$\\
$\beta$-OmniDecVAE & $\mathbf{0.054 \pm 0.001}$ & $\mathbf{0.153 \pm 0.002}$ & $0.237 \pm 0.005$ & $\mathbf{0.078 \pm 0.004}$ & $0.093 \pm 0.006$ & $2.700 \pm 0.220$\\
\midrule\multicolumn{7}{l}{\textit{UCI HAR ($C=9$)}}\\
MMVAE & $0.212 \pm 0.000$ & $0.290 \pm 0.000$ & $0.643 \pm 0.007$ & $\mathbf{0.141 \pm 0.010}$ & $0.148 \pm 0.001$ & $3.038 \pm 0.082$\\
$\beta$-MMVAE & $0.191 \pm 0.000$ & $\mathbf{0.248 \pm 0.000}$ & $1.671 \pm 0.028$ & $\underline{0.150 \pm 0.019}$ & $0.136 \pm 0.002$ & $2.984 \pm 0.057$\\
OmniDecVAE & $0.226 \pm 0.003$ & $0.374 \pm 0.003$ & $\mathbf{0.249 \pm 0.006}$ & $0.541 \pm 0.013$ & $0.074 \pm 0.004$ & $3.719 \pm 0.208$\\
$\beta$-OmniDecVAE & $\mathbf{0.127 \pm 0.002}$ & $0.264 \pm 0.002$ & $0.389 \pm 0.035$ & $0.169 \pm 0.008$ & $\mathbf{0.050 \pm 0.006}$ & $\mathbf{2.617 \pm 0.129}$\\
\midrule\multicolumn{7}{l}{\textit{WESAD ($C=14$)}}\\
MMVAE & $0.230 \pm 0.000$ & $0.243 \pm 0.000$ & $0.436 \pm 0.007$ & $0.203 \pm 0.009$ & $0.202 \pm 0.001$ & $2.232 \pm 0.055$\\
$\beta$-MMVAE & $0.218 \pm 0.000$ & $0.214 \pm 0.000$ & $1.025 \pm 0.012$ & $\mathbf{0.164 \pm 0.006}$ & $0.197 \pm 0.000$ & $2.258 \pm 0.037$\\
OmniDecVAE & $0.113 \pm 0.001$ & $0.238 \pm 0.001$ & $\mathbf{0.198 \pm 0.002}$ & $0.425 \pm 0.003$ & $\underline{0.077 \pm 0.011}$ & $1.803 \pm 0.427$\\
$\beta$-OmniDecVAE & $\mathbf{0.070 \pm 0.000}$ & $\mathbf{0.182 \pm 0.001}$ & $0.345 \pm 0.005$ & $\underline{0.168 \pm 0.002}$ & $\mathbf{0.066 \pm 0.012}$ & $\mathbf{1.160 \pm 0.293}$\\
\bottomrule\end{tabular}}\end{adjustbox}
\par\smallskip\raggedright\scriptsize $n=5$ pre-training seeds. Mean $\pm$ 95\% CI ($t$-distribution) reported. Bold: best per column. Underlined: not significantly different from the best (Welch $t$-test, $p\geq0.05$).
\end{table*}

Table \ref{tab:generation_quality} reports generative performance in the SI splits. On HARWE, the OmniDecVAE models reconstruct the test set more faithfully (lower MSE and MAE) and have lower latent distances ($d_{L2}$ and $d_{cos}$) between real and reconstructed embeddings; the MMVAE models generate from latent noise closer to the data distribution (lower MK-MMD and DivScore). On UCI HAR and WESAD, $\beta$-OmniDecVAE$_{Enc-Dec}$ has the lowest MSE and MAE on WESAD, while MMVAE has the lowest MAE on UCI HAR), and generates closer to the data distribution (lower MK-MMD and DivScore). OmniDecVAEs thus provide the most faithful component-wise reconstruction on all three datasets, and better unconditional generation on UCI HAR and WESAD, though not on HARWE.

\begin{figure*}[!t]\centering
\includegraphics[width=\textwidth]{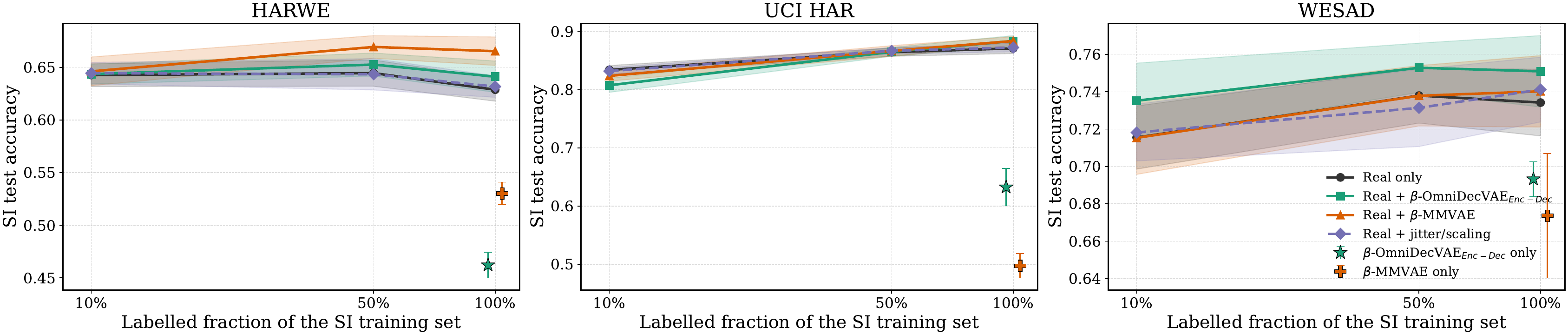}
\caption{\textcolor{black}{Utility of generated data in the SI splits of HARWE~\cite{Esmaeilzehi2024HARWE:Environments} (HAR), UCI HAR~\cite{anguita2013public} (HAR), WESAD~\cite{schmidt2018introducing} (AR): mean accuracy and CI of an MLP trained on a labeled fraction of the real data, alone or augmented with synthetic or jittered and scaled frames, or on generated data only (markers at 100\%, mean $\pm$ 95\% CI over 25 seeds).}}
\label{fig:utility_generated}
\end{figure*}

In Fig~\ref{fig:utility_generated} we examine the utility of synthetic samples as training data. On WESAD, augmentation with $\beta$-OmniDecVAE$_{Enc-Dec}$ improves accuracy at every labeled fraction, whereas $\beta$-MMVAE does not; on UCI HAR, both generators improve accuracy with the full training set but not when $10\%$ of it is used. On HARWE, $\beta$-MMVAE augmentation is more effective than $\beta$-OmniDecVAE$_{Enc-Dec}$ as the labeled fraction increases, and both outperform conventional augmentations. Moreover, classifiers trained on synthetic data only are far less accurate than those trained on the full real set. Therefore, synthetic data modestly improves SI recognition, depending on the dataset and the amount of labeled data. 

\subsection{Ablation Studies}

\subsubsection{Contribution of Multi-modal Input}
\label{modalities_ablation}
Tables \ref{tab:modalities_ablation_activity},\ref{tab:modalities_ablation_identity} presents downstream HAR and IR performance for different combinations of modalities. CNN-Transformer is the strongest model with only audio input, as its YAMNet~\cite{yamnet2019} backbone is designed for audio. For other modality subsets, $\beta$-OmniDecVAE$_{Enc-Dec}$ and $\beta$-MMVAE are comparable, each leading some comparisons. In the SD scenario, inertial sensor subsets outperform the full set ($C=30$) indicating their importance for HAR with known subjects, whereas in the SI scenario the full set provides the best results overall.

\begin{table*}[ht!]
\centering
\caption{Ablation Study on Number of Modalities \textcolor{black}{on HARWE~\cite{Esmaeilzehi2024HARWE:Environments}} - Activity Recognition Performance}
\scriptsize
\scriptsize
\resizebox{\textwidth}{!}{%
\begin{tabular}{c|cc|cc|cc|cc|cc|cc}
\toprule   
\multirow{3}{*}{Modalities} & 
\multicolumn{6}{c|}{SD HAR \textuparrow} &
\multicolumn{6}{c}{SI HAR \textuparrow}
\\ 
\cmidrule(lr){2-13} 
 & \multicolumn{2}{c|}{CNN-Transformer \cite{Esmaeilzehi2024HARWE:Environments} \textuparrow} & \multicolumn{2}{c|}{$\beta$-MMVAE \textuparrow} &
\multicolumn{2}{c|}{$\beta$-OmniDecVAE$_{Enc-Dec}$ \textuparrow} & \multicolumn{2}{c|}{CNN-Transformer \cite{Esmaeilzehi2024HARWE:Environments} \textuparrow} & \multicolumn{2}{c|}{$\beta$-MMVAE \textuparrow} & \multicolumn{2}{c}{$\beta$-OmniDecVAE$_{Enc-Dec}$ \textuparrow}\\

\cmidrule(lr){2-13} 
 & Acc. \textuparrow & F1 \textuparrow & Acc. \textuparrow & F1 \textuparrow & Acc. \textuparrow & F1 \textuparrow  & Acc. \textuparrow & F1 \textuparrow & Acc. \textuparrow & F1 \textuparrow & Acc. \textuparrow & F1 \textuparrow  \\

\midrule
\midrule
Smartwatch (BVP+EDA+TEMP, $C=3$) & 17.87 $\pm$ 1.20 & 16.23 $\pm$ 1.23 & \textbf{22.50} $\pm$ 0.15 & \textbf{21.99} $\pm$ 0.16 & 21.68 $\pm$ 0.20 & 21.51 $\pm$ 0.18 & 11.24$\pm$ 1.03 & 8.69 $\pm$ 1.07 & \textbf{13.70} $\pm$ 0.19  & \textbf{13.58} $\pm$ 0.35 & 12.78 $\pm$ 0.17 & 12.63 $\pm$ 0.15 \\

Audio (6 OCs, $C=6$) & \textbf{48.30} $\pm$ 9.26 & \textbf{45.87} $\pm$ 10.89 & 38.80 $\pm$ 0.19 & 37.93 $\pm$ 0.17 & 40.85 $\pm$ 0.28 & 39.75 $\pm$ 0.30 & \textbf{41.61} $\pm$ 8.31 & \textbf{38.48} $\pm$ 9.94 & 35.62 $\pm$ 0.20 & 33.80 $\pm$ 0.20 & 38.07 $\pm$ 0.16 & 36.46 $\pm$ 0.15\\

SmartWatch + Audio ($C=9$) & 43.55 $\pm$ 4.92 & 40.37 $\pm$ 5.43 & 47.33 $\pm$  0.08 & 46.58 $\pm$ 0.11 & \textbf{49.62} $\pm$ 0.09 & \textbf{48.91} $\pm$ 0.11 & \textbf{38.75} $\pm$ 5.42 & 36.78 $\pm$ 5.54 & 37.81 $\pm$ 0.31 & 36.46 $\pm$ 0.32 & 38.33 $\pm$ 0.22 & \textbf{37.02} $\pm$ 0.23 \\

Inertial ($C=21$) & 58.35 $\pm$  4.26 & 55.92 $\pm$ 4.46 & 86.80 $\pm$ 0.15 & \textbf{86.80} $\pm$ 0.15 & \textbf{86.24} $\pm$ 0.11 & 86.23 $\pm$ 0.12 & 32.80 $\pm$ 2.77 & 26.75 $\pm$ 3.25 & 57.91\textbf{} $\pm$ 0.06 & 56.90 $\pm$ 0.09 & 57.16 $\pm$ 0.11 & \textbf{56.93}  $\pm$ 0.09  \\

Inertial + Smartwatch ($C=24$) & 49.85 $\pm$ 3.56 & 46.63 $\pm$ 4.20 & 86.00 $\pm$ 0.15 & \textbf{86.23}  $\pm$ 0.15 & \textbf{86.20}  $\pm$ 0.17 & 86.19 $\pm$ 0.16 & 32.46$\pm$ 2.20 & 27.09 $\pm$ 2.70 & \textbf{62.30} $\pm$ 0.12 & \textbf{62.11} $\pm$ 0.10 & 55.13 $\pm$ 0.21 & 54.81 $\pm$ 0.20 \\

Inertial + Audio ($C=27$) & 40.63 $\pm$ 4.67 & 36.05 $\pm$ 4.07 & \textbf{87.21} $\pm$ 0.07 & \textbf{87.18} $\pm$ 0.08 & 81.13 $\pm$ 0.14  & 81.04 $\pm$ 0.14 & 36.02 $\pm$ 2.95 & 31.12 $\pm$ 3.65 & \textbf{69.42} $\pm$ 0.16 & \textbf{68.59} $\pm$ 0.17 & 62.63 $\pm$ 0.12 & 62.00 $\pm$ 0.10 \\

Smartwatch + Inertial + Audio ($C=30$) & 39.93 $\pm$ 6.03 & 35.70 $\pm$ 6.65 & 82.36 $\pm$ 0.13 & 82.31 $\pm$ 0.13 & \textbf{84.56} $\pm$ 0.21 & \textbf{84.48} $\pm$ 0.21 & 32.98 $\pm$ 2.32 & 28.30 $\pm$ 2.72 & \textbf{70.91} $\pm$ 0.10  & \textbf{70.37} $\pm$ 0.10 & 64.56 $\pm$ 0.12 & 64.09 $\pm$ 0.14 \\

\bottomrule
\end{tabular}
}%
\par\smallskip\scriptsize $n=25$ ($5$ pre-training $\times5$ classifier seeds), mean $\pm95\%$ CIs reported. Acc.: Accuracy. Best Acc. and F1 score for each row of each task is given in \textbf{bold}.
\label{tab:modalities_ablation_activity}
\end{table*}

\begin{table*}[ht!]
\centering
\caption{Ablation Study on Number of Modalities \textcolor{black}{on HARWE~\cite{Esmaeilzehi2024HARWE:Environments}} - Identity Recognition Performance}
\scriptsize
\scriptsize
\resizebox{\textwidth}{!}{%
\begin{tabular}{c|cc|cc|cc}
\toprule   
\multirow{3}{*}{Modalities} & 
\multicolumn{6}{|c}{SD IR \textuparrow} 
\\ 
\cmidrule(lr){2-7} 
 & \multicolumn{2}{c|}{CNN-Transformer \cite{Esmaeilzehi2024HARWE:Environments} \textuparrow} & \multicolumn{2}{c|}{$\beta$-MMVAE \textuparrow} &
\multicolumn{2}{c}{$\beta$-OmniDecVAE$_{Enc-Dec}$ \textuparrow} \\

\cmidrule(lr){2-7} 
 & Acc. \textuparrow & F1 \textuparrow & Acc. \textuparrow & F1 \textuparrow & Acc. \textuparrow & F1 \textuparrow  \\

\midrule
\midrule
Smartwatch (BVP+EDA+TEMP, $C=3$) & 38.57 $\pm$ 4.74 & 34.88 $\pm$ 4.92 & \textbf{50.50} $\pm$ 0.14 & \textbf{48.45} $\pm$ 0.13 & 45.50 $\pm$ 0.13 & 44.02 $\pm$ 0.09 \\

Audio (6 OCs, $C=6$) &  3.35 $\pm$ 0.002 & 0.22 $\pm$ 0.001 & \textbf{17.56} $\pm$ 0.29 & \textbf{16.29} $\pm$ 0.29 & 16.75 $\pm$ 0.28 & 15.24 $\pm$ 0.28 \\

SmartWatch + Audio ($C=9$) & 33.21 $\pm$ 4.35 & 29.26 $\pm$ 4.66  &  47.04 $\pm$ 0.29 & 46.68 $\pm$ 0.28 & \textbf{48.27} $\pm$ 0.15 & \textbf{47.59} $\pm$ 0.17 \\

Inertial ($C=21$) &  74.04 $\pm$ 1.23 & 74.59 $\pm$ 1.21 & \textbf{88.24} $\pm$ 0.25  &\textbf{88.21} $\pm$ 0.26 & 88.01 $\pm$ 0.12 & 87.96 $\pm$ 0.12 \\

Inertial + Smartwatch ($C=24$)  & 62.77 $\pm$ 4.05 & 61.78 $\pm$ 4.38 & \textbf{92.77} $\pm$ 0.11  & \textbf{92.76} $\pm$ 0.10 & 92.66 $\pm$ 0.06  & 92.63 $\pm$ 0.06 \\

Inertial + Audio ($C=27$) &  36.29 $\pm$ 4.17 & 35.41 $\pm$ 4.21 & \textbf{85.66} $\pm$ 0.12 & \textbf{85.68} $\pm$ 0.12 & 75.63 $\pm$ 0.16 & 75.63 $\pm$ 0.16 \\

Smartwatch + Inertial + Audio ($C=30$) & 67.36 $\pm$ 2.92 & 66.85 $\pm$ 3.05 &  \textbf{79.42} $\pm$ 0.34 & 79.65 $\pm$ 0.35 & \textbf{88.97} $\pm$ 0.13 & 88.99 $\pm$ 0.13 \\

\bottomrule
\end{tabular}
}%
\par\smallskip\scriptsize
$n=25$ ($5$ pre-training $\times5$ classifier seeds), mean $\pm95\%$ CIs reported. Acc.: Accuracy. Best Acc. and F1 score for each row of each task is given in \textbf{bold}.
\label{tab:modalities_ablation_identity}
\end{table*}

\subsubsection{Contribution of Optimization Terms}
\label{loss_ablation}
Table \ref{tab:loss_ablation} contains a detailed ablation of the four terms contained in Eq.~\eqref{eq_mmvda_dec}; in terms of task-related performance, we see that utilizing the $\mathcal{L}_{SSLDec}$ terms alongside a supervised loss, gives the highest performance while maintaining disentanglement between activity and identity. 
It becomes evident that $\mathcal{L}_{SSLDec}$ and the decoder are the main drivers of disentanglement, as their absence results in a larger imbalance in the performance between HAR and IR. \textcolor{black}{Measured with $D$ on HARWE (SD), most of the disentanglement of these models indeed comes from the decoder and $\mathcal{L}_{SSLDec}$ ($D=23.0$ with the decoder, $15.3$ with $\mathcal{L}_{SSLDec}$, $24.2$ with both).}

\begin{table}[ht!]
\centering
\caption{Ablation Study on OmniDecVAE Loss Components}
\scriptsize
\scriptsize
\resizebox{\columnwidth}{!}{%
\begin{tabular}{c|cc|cc}
\toprule   
\multirow{2}{*}{Loss Components} & \multicolumn{2}{|c}{SD HAR \textuparrow} &
\multicolumn{2}{|c}{SD IR \textuparrow} \\
\cmidrule{2-5}
 & Acc. \textuparrow & F1 \textuparrow & Acc \textuparrow  & F1 \textuparrow\\

\midrule
\midrule
$\mathcal{L}_{SSLDec}$ & 76.03 & 75.91 & 72.37 & 72.42 \\
$\mathcal{L}_{dec}$ & 83.22 & 83.14 & 86.92 & 86.93 \\
$\mathcal{L}_{prior}$ & 65.30 & 64.97  & 47.69 & 48.09   \\
Supervised & 82.07 & 82.00 & 83.21 & 83.23   \\
$\mathcal{L}_{SSLDec}$ + $\mathcal{L}_{prior}$  & 74.66 & 74.56 & 67.26 & 67.31 \\
$\mathcal{L}_{SSLDec}$ + $\mathcal{L}_{dec}$  & 82.09 & 82.01 & 86.58 & 86.63  \\
$\mathbf{\mathcal{L}_{SSLDec}}$ \textbf{+ Supervised} & \textbf{85.83} & \textbf{85.75} & \textbf{88.32} & \textbf{88.33}  \\
$\mathcal{L}_{dec}$ + $\mathcal{L}_{prior}$ & 81.31 & 81.23 & 85.70 & 85.69 \\
$\mathcal{L}_{dec}$ + Supervised  & 83.06 & 82.96 & 86.46 & 86.49 \\
$\mathcal{L}_{prior}$ + Supervised  & 70.64 & 70.44 & 56.46 & 56.93 \\
$\mathcal{L}_{SSLDec}$ + $\mathcal{L}_{prior}$ + $\mathcal{L}_{dec}$ & 82.17 & 82.04 & 85.89 & 85.89 \\
$\mathcal{L}_{SSLDec}$ + $\mathcal{L}_{prior}$ + Supervised & 72.48 & 72.43 & 62.28 & 62.45 \\
$\mathcal{L}_{SSLDec}$ + $\mathcal{L}_{dec}$ + Supervised & 83.80 & 83.71 & 87.71 & 87.73\\
$\mathcal{L}_{dec}$ + $\mathcal{L}_{prior}$ + Supervised & 82.11 & 82.02 & 85.35 & 85.35 \\
$\mathcal{L}_{SSLDec}$ + $\mathcal{L}_{prior}$ + $\mathcal{L}_{dec}$ + Supervised & 81.89 & 81.78 & 86.39 & 86.38 \\
\bottomrule
\end{tabular}
}%
\label{tab:loss_ablation}
\end{table}

\subsubsection{Contribution of Gaussian Prior Approximation Weight $\beta$}
\label{beta_ablation}
\begin{figure*}[ht!]
    \includegraphics[height=5.5cm, width=0.8\textwidth]{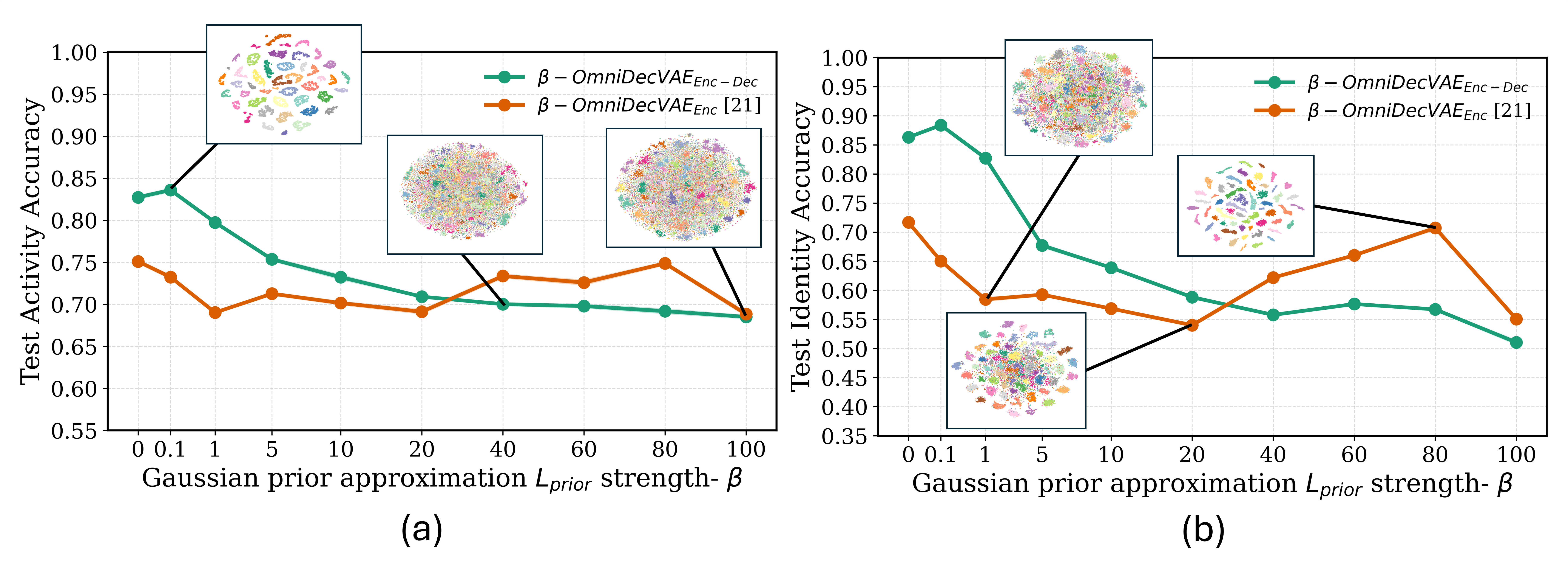}
    \centering
    \caption{The effect of varying Gaussian prior approximation strength $\beta$ on representation disentanglement for (a) HAR and (b) IR for $\beta$-OmniDecVAE$_{Enc}$ (orange) and $\beta$-OmniDecVAE$_{Enc-Dec}$ (green) \textcolor{black}{on the HARWE~\cite{Esmaeilzehi2024HARWE:Environments} dataset}.}
    \label{fig:beta_ablation}
\end{figure*}

The role of $\mathcal{L}_{prior}$ and the value of $\beta$ is better understood through Fig.~\ref{fig:beta_ablation}. As discussed in DecVAEs~\cite{Ziogas2026VariationalRepresentations}, a strong $\beta$ compression strength may not always result in a better representation; in Fig.~\ref{fig:beta_ablation}a, HAR performance correlates directly with modality disentanglement at different values of $\beta$. Here higher $\beta$ values result in a collapse of the disentangled structure. We also see that the absence of the $\mathcal{L}_{prior}$ through $\beta=0$ is not detrimental to the downstream performance, which we attribute to the disentanglement mechanisms of $\mathcal{L}_{SSLDec}$. A different behavior is evident in the decoder-less variant of OmniDecVAE (see Fig.~\ref{fig:beta_ablation}b); in the absence of a decoder, a higher $\beta$ is needed for a disentangled representation. 

\subsubsection{Multi-modal Anchor Construction}
\label{ablation_anchor_results}

\begin{figure*}[!t]\centering
\includegraphics[width=\textwidth]{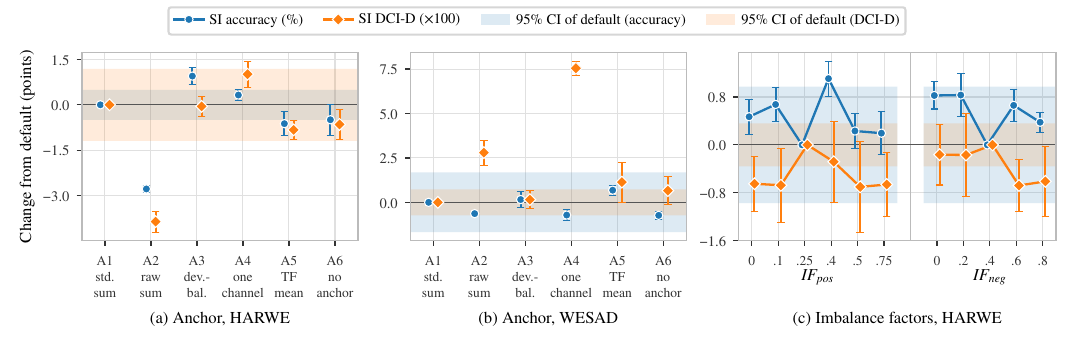}
\caption{\textcolor{black}{Anchor and asymmetric contrastive weighting ablations in $\beta$-OmniDecVAE$_{Enc-Dec}$, SI split of HARWE~\cite{Esmaeilzehi2024HARWE:Environments}: change from the default standardized sum (A1 at $0.0$) in HAR accuracy (\%) and DCI D. (a,b) Different anchor construction options, (c) Imbalance factors on HARWE, varied one at a time (defaults $IF_{pos}=0.25$, $IF_{neg}=0.4$). Shaded bands: 95\% CI of the default over 3 pre-training seeds. Other settings: 95\% CI over 5 classifier states (accuracy) or 5 cross-validation folds (D).}}
\label{fig:anchor_weights_ablation}
\end{figure*}

We compare the default anchor construction scheme $A(\cdot)$ (sum of standardized channels (A1, Eq.~\eqref{eq_anchorsum}) with: the sum of raw channels (A2); a device-balanced sum where each channel is scaled by the inverse square root of the components of that device (A3); a specific channel as the anchor - the same channel is kept as a single component (audio in HARWE and chest EMG in WESAD, as the most expressive modalities) (A4); the mean of components' TF features (A5); no anchor in optimization and evaluation (A6). We pretrain $\beta$-OmniDecVAE$_{Enc}$, $\beta$-OmniDecVAE$_{Enc-Dec}$ on the HARWE and WESAD SI splits. In Fig.~\ref{fig:anchor_weights_ablation}a,b, anchors of $\beta$-OmniDecVAE$_{Enc-Dec}$ perform similarly on HARWE except A2, which is significantly worse, perhaps due to the different units of the inputs. On WESAD, A2 and A4 yield increased DCI. These results highlight A1 as the simplest, not necessarily best performing, robust choice, as A4 requires selecting a channel per dataset. Without an anchor (A6), accuracy drops slightly on both datasets and D increases slightly on WESAD; as anticipated in Section~\ref{anchor_methods}, the anchor adds little when a decoder is present. Without a decoder ($\beta$-OmniDecVAE$_{Enc}$), the removal of the anchor significantly lowers DCI D on WESAD.

\subsubsection{\textcolor{black}{Asymmetric Contrastive Weighting}}
\label{ablation_weights_results}
\textcolor{black}{We ablate $IF_{pos}\in \{0, 0.1, 0.25, 0.4, 0.5, 0.75\}$ and $IF_{neg}\in \{0, 0.2, 0.4, 0.6, 0.8 \}$ one at a time, and pre-train OmniDecVAEs on the SI splits on the three datasets, where $IF=0$ implies uniform weights. UCI HAR is evaluated for the $IF_{neg}$ experiment only, as $IF_{pos}=0$ due to the absence of a stronger component, whereas in WESAD we use the chest EMG as the stronger component that requires $IF_{pos}>0$. Fig.~\ref{fig:anchor_weights_ablation}c presents the sensitivity of MMVDA to the two imbalance factors varying values on HARWE. For $\beta$-OmniDecVAE$_{Enc-Dec}$, we see a small non-significant variation across different imbalance values on HARWE, indicating that the model is robust to the imbalance factors values.}

\subsection{Complexity}
\label{complexity}

\begin{figure*}[h!]
    \includegraphics[height = 5.5cm,width=0.8\textwidth]{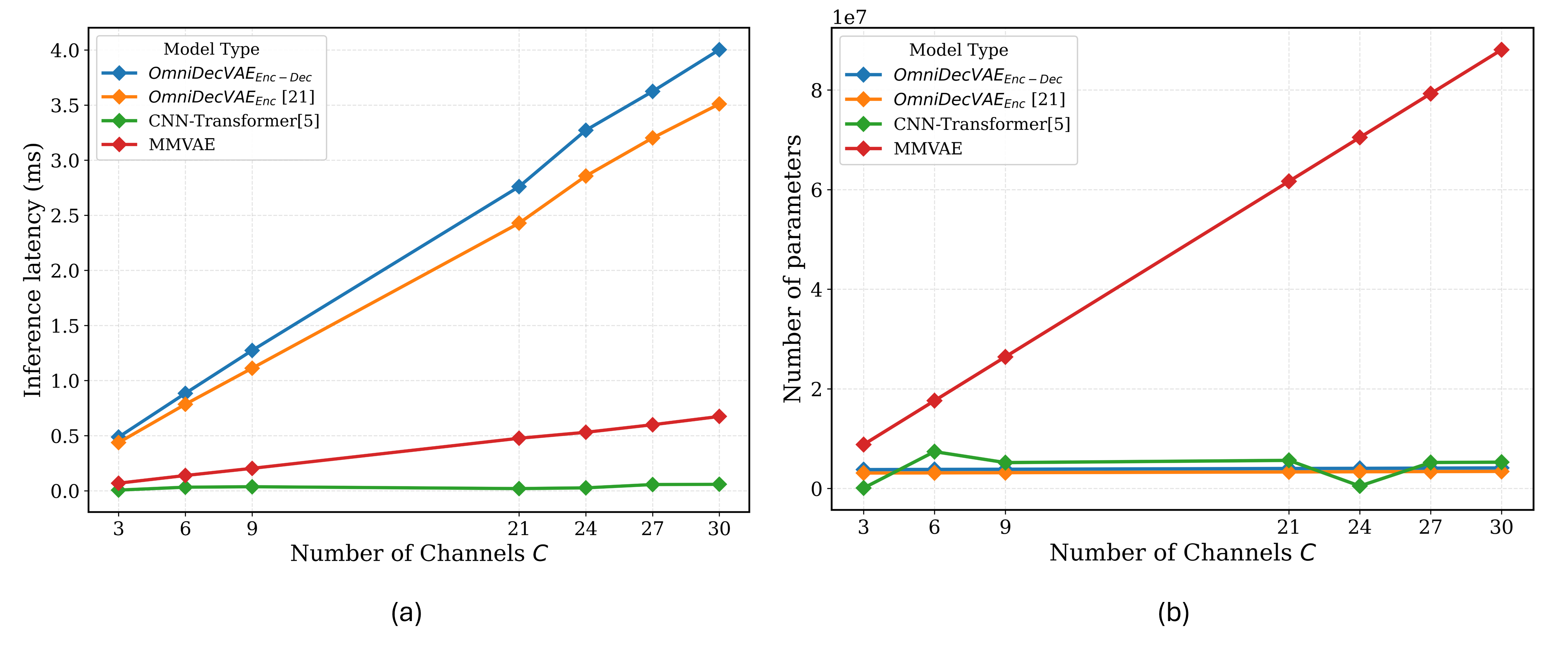}
    \centering
    \caption{The effect of a varying number of input channels on (a) inference latency and (b) number of parameters.}
    \label{fig:complexity}
\end{figure*}

In Table \ref{tab:complexity}, we evaluate our proposed OmniDecVAE alongside a Transformer-based model and MMVAE on their complexity, by calculating per-sample FLOPs and latency inside a batched input \textcolor{black}{on the GPU used for training}. The CNN-Transformer of \cite{Esmaeilzehi2024HARWE:Environments} is the fastest model with very low latency per sample, and MMVAE performs the fewest FLOPs. OmniDecVAE is the lightest alternative in terms of SoD and number of parameters, but it requires the most operations per sample and has a higher latency \textcolor{black}{as a single shared encoder is used to encode $C+1$ inputs}; at the same time, MMVAE presents increasing storage requirements due to the high number of modalities it accommodates as separate branches. Indeed, Fig.~\ref{fig:complexity}b showcases that the spatial complexity of MMVAEs increase with the number of modalities; on the other hand OmniDecVAEs are invariant to the number of modalities for that matter, while being more prone to fast inference as the number increases. 

\begin{table}[ht!]
\centering
\caption{Performance on Complexity Metrics for $C=30$ Modality Channels}

\resizebox{\columnwidth}{!}{%
\begin{tabular}{c|cccc}
\toprule   
Model & GFLOPs \textdownarrow &
Latency$_{Inf}$ (ms) \textdownarrow & SoD (MB) \textdownarrow& Parameters (M) \textdownarrow \\
\midrule
\midrule
CNN-Transformer \cite{Esmaeilzehi2024HARWE:Environments} & 0.65 & \textbf{0.06} & 20.19 & 5.28 \\
MMVAE & \textbf{0.45} & 0.67 & 336.46 & 88.09 \\
OmniDecVAE$_{Enc}$ \cite{Ziogas2026VariationalRepresentations}
& 29.61 & 3.51 & \textbf{13.20} & \textbf{3.43}\\
OmniDecVAE$_{Enc-Dec}$ 
& 85.85 & 4.00 & 15.82 & 4.11 \\
\bottomrule
\end{tabular}
}%
\par\smallskip\scriptsize Latency$_{Inf}$: Inference latency, SoD: size on disk
\label{tab:complexity}
\end{table}

\subsection{Discussion}
\label{discussion}

Wearable representation learning has significantly advanced in the latest years mainly through fusion architectures that integrate multi-modal streams across large-scale datasets, spearheaded by the prowess of transformer-based models. However, despite high HAR classification performance, the focus of these models does not extend to full-stack representations. Here, we presented omni-modal representation learning for wearable computing, by efficiently learning through a modality-invariant architecture; modality-specific data handling for OmniDecVAEs ends at the choice of input representation and fusion is achieved through the MMVDA learning objective rather than architectural interventions. \textcolor{black}{On HARWE,} OmniDecVAE does not utilize the video modality, one of the most expressive modalities for HAR, and yet performs on par \textcolor{black}{(Acc. of $64.6\%$ SI HAR)} with a supervised transformer-based method in the HARWE SI scenario that utilizes video (Acc. of $68.5\%$ in SI HAR) \cite{Esmaeilzehi2024HARWE:Environments}. 

\textcolor{black}{Notably, in quantitative disentanglement results, $\beta$-OmniDecVAE$_{Enc-Dec}$ consistently learns more disentangled representations than baselines across HARWE, UCI HAR, and WESAD, without compromising their informativeness compared to VAE baselines. The encoder-only version of OmniDecVAE was found to be more robust to confounding interventions to factors (IRS), and its latent dimensions learned have lower statistical interdependence (GCN) than competitors. Our ablations on the MMVDA objective loss terms attribute the disentanglement properties to a combination of the $\mathcal{L}_{SSLDec}$ objective, and importantly, the decoder in OmniDecVAE$_{Enc-Dec}$ (Table~\ref{tab:loss_ablation}), whereas the value of $\beta$ regularization of $\mathcal{L}_{prior}$ is also decisive for disentanglement and performance (Fig.~\ref{fig:beta_ablation}).}
Visual inspection of the representations in Fig.~\ref{fig:tsne} reveals well delimited modes that arise as a result of subspace learning through MMVDA. \textcolor{black}{Disentanglement fosters downstream recognition but also depends on the composition of the data; OmniDecVAEs is the most accurate model in the SD scenario of HARWE in both HAR and IR, which resembles a biometric identification setting where each subject is enrolled beforehand. On WESAD SD scenario, OmniDecVAE trails supervised late-fusion CNN in AR and IR, but performs better than VAE-based baselines and the supervised CNN-Transformer. On unseen subjects in the two multi-device datasets, OmniDecVAE outperforms by a large margin the supervised models($64.6\%$ vs. at most $33.0\%$ on HARWE, $74.6\%$ vs. at most $52.3\%$ on WESAD). On the single-device UCI HAR, end-to-end supervised models perform clearly more accurate in both SD and SI.} 

\textcolor{black}{
In multi-modal synthetic data generation, OmniDecVAE reconstructs more faithfully than the multi-branch MMVAEs on all three datasets, using a single shared decoder. Unconditional generation from latent noise is closer to the real distribution in UCI HAR and WESAD, but not on HARWE, and its utility as a wearable omni-modal time series generator improves SI AR on WESAD when synthetic samples are used as augmentations, with modest gains across datasets. Synthetic data showed a beneficial effect as a complementary training mechanism but not as a standalone source of information. Finally, the size of OmniDecVAEs is invariant to the number of modalities and has a minimal storage footprint (as low as $13.2MB$), an advantage over architectures with modality-specific branches, at the cost of more operations per sample (see Section~\ref{complexity}).}

\textcolor{black}{Overall, OmniDecVAE is not the best model in all settings, as multi-branch supervised transformer and VAE-based architectures keep an edge in specific settings. $\beta$-MMVAE is more accurate for SI HAR on HARWE ($70.9\%$ vs. $64.6\%$) and on UCI HAR ($81.6\%$ vs. $79.8\%$); on HARWE, this SI performance gap appears only when inertial channels are combined with other modalities (Table~\ref{tab:modalities_ablation_activity}). Because OmniDecVAE has a shared encoder, one set of parameters serves all components, which possibly leads to increased superposition of features on a fixed representational capacity, whereas in all baselines, dedicated branches per modality are allocated; conditioning the encoder on the component, as done for the decoder (Section~\ref{decoder_theory}) could be a potential fix to minimize the influence of this bottleneck. Moreover, supervised end-to-end models are more accurate when trained on the single-device UCI HAR, or on the SD split of WESAD. These results hint that the proposed self-supervised omni-modal framework is most valuable when the components have a high heterogeneity and number, and when unseen subjects are evaluated. For generation, OmniDecVAE reconstructs more faithfully on all three datasets, but on HARWE its latent noise generation is less close to the real data distribution than MMVAE, although the reverse holds for UCI HAR and WESAD; the utility of OmniDecVAE synthetic data was found to be modest and dataset-dependent (Fig.~\ref{fig:utility_generated}).}

\textcolor{black}{These properties are relevant to healthcare applications, such as monitoring with missing or changing sensors in-the-wild and in clinical environments, identity-aware modeling and privacy, and edge deployment, although these uses were not evaluated here and remain as future applications.} \textcolor{black}{Along these lines, other aspects for future work remain.} Our implementation is based on TF representations; although STFT and Mel scale are more informative and predominantly used in deep learning, the generation of synthetic data in the time domain would allow more freedom in downstream processing. Moreover, we did not experiment with cross-modal inference or missing modality scenarios, for which the omni-modal separation and conservation of hierarchical relations within the latent representations of OmniDecVAEs, hints a promising starting point. Future extensions of our work will also accommodate modalities beyond time series, such as video or text.

\section{Conclusion} 
\label{conclusion}
This work proposes OmniDecVAEs, a structured representation learning framework that learns full-stack wearable disentangled representations by fusing arbitrarily large multi-modal (or omni-modal) data streams with a single shared encoder-decoder scheme. \textcolor{black}{Our approach extends variational decomposition autoencoding with a multi-modal anchor, an asymmetric decoder design, and weighting of modalities, thus enabling a modality-invariant learnable objective that controls fusion, alignment and separation of different channels.} Our evaluations highlight full-stack representational capacity \textcolor{black}{of OmniDecVAEs;} \textcolor{black}{on the omni-modal HARWE dataset with up to thirty channels and the two public datasets UCI HAR and WESAD, OmniDecVAEs learn the most disentangled representations on unseen subjects (DCI D improves by up to $3$ percentage points (pp)), reconstruct more faithfully than multi-modal VAEs ($34\%$ to $74\%$ lower MSE), and are the most accurate in the subject-dependent setting (HARWE HAR and IR up by $1.1$pp and $5.8$pp) and more accurate than end-to-end supervised models for unseen subjects in two datasets (HARWE $+31.6$pp, WESAD $+22.3$pp). They do so with modality-invariant storage requirements of $4.1M$ parameters, though at the cost of more operations per sample. Their synthetic data improve affect recognition on unseen WESAD subjects by $1.5-2.0$ pp, whereas on other datasets the quality of synthetic data varies. They are also less accurate than a multi-branched VAE for unseen subjects on HARWE (by $-6.3$pp), and than supervised end-to-end models on the inertial-only UCI HAR and known subjects of WESAD.} These results support OmniDecVAEs as a scalable approach for next-generation full-stack models in wearable AI; \textcolor{black}{their use in real world clinical applications, biometric security, and missing data synthesis, remains to be validated.}



\section*{References}

\bibliographystyle{ieeetr}
\bibliography{references}

\end{document}